\documentclass[11pt]{article}

\usepackage[preprint]{acl}

\usepackage{times}
\usepackage{latexsym}
\usepackage[table]{xcolor}
\usepackage{amssymb}
\usepackage{listings}
\usepackage{fvextra} 
\usepackage{longtable}
\usepackage{booktabs}
\DefineVerbatimEnvironment{CodeBlock}{Verbatim}{
  breaklines=true,
  breakanywhere=true,    
  fontsize=\footnotesize,
  commandchars=\\\{\},   
  gobble=0
}
\usepackage{amsmath} 
\usepackage[T1]{fontenc}

\usepackage[utf8]{inputenc}

\usepackage{microtype}

\usepackage{inconsolata}

\usepackage{graphicx}
\usepackage{xspace}
\usepackage{enumitem}
\usepackage{pifont}
\usepackage[most]{tcolorbox}
\usepackage{lipsum}
\usepackage{xcolor}
\usepackage{cuted}
\usepackage{booktabs}
\usepackage{multirow}
\tcbuselibrary{listings}
\usepackage{pifont}
\newcommand{\cmark}{\ding{51}}
\newcommand{\xmark}{\ding{55}}

\newcommand{\circone}{\ding{172}\xspace}
\newcommand{\circtwo}{\ding{173}\xspace}
\newcommand{\circthree}{\ding{174}\xspace}

\newcommand{\zz}[1]{{\color{magenta}[ZZ: {#1}]}}

\newcommand{\tabref}[1]{Table~\ref{#1}}
\newcommand{\figref}[1]{Fig.~\ref{#1}}

\newcommand{\secref}[1]{\S\ref{#1}}
\newcommand{\appref}[1]{Appendix~\ref{#1}}

\newcommand{\algname}{\textsc{TARo}\xspace}

\definecolor{promptborder}{RGB}{60,90,130}
\definecolor{promptbg}{RGB}{247,249,252}
\definecolor{prompttitlebg}{RGB}{233,238,245}
\definecolor{prompttitlefg}{RGB}{25,35,55}

\tcbset{
  promptbox/.style={
    enhanced,
    breakable,
    colback=promptbg,
    colframe=promptborder,
    boxrule=0.6pt,
    arc=1.5mm,
    left=1.2mm,
    right=1.2mm,
    top=1mm,
    bottom=1mm,
    sharp corners=south,
    coltitle=prompttitlefg,
    fonttitle=\bfseries,
    title filled=true,
    colbacktitle=prompttitlebg,
    attach boxed title to top left={
      xshift=2mm,
      yshift*=-\tcboxedtitleheight/2
    },
    boxed title style={
      boxrule=0pt,
      frame hidden,
      arc=1mm,
      left=1mm,
      right=1mm,
      top=0.6mm,
      bottom=0.6mm
    },
    borderline west={1.5pt}{0pt}{promptborder},
    before skip=10pt,
    after skip=10pt
  }
}

\title{TARo: Token-level Adaptive Routing for LLM Test-time Alignment}

\author{
Arushi Rai$^{*~1~2}$ \quad 
Qiang Zhang$^1$ \quad 
Hanqing Zeng$^1$ \quad 
Yunkai Zhang$^3$ \\[0.3em]
\textbf{Dipesh Tamboli}$^1$ \quad 
\textbf{Xiangjun Fan}$^1$ \quad 
\textbf{Zhuokai Zhao}$^{\dagger~1}$ \quad
\textbf{Lizhu Zhang}$^{\dagger~1}$  \\[0.3em]
*Work done during internship at Meta. \\
$^\dagger$ Joint last author   \\[0.3em]
$^1$Meta \quad 
$^2$University of Pittsburgh \quad
$^3$University of California, Berkeley \\[0.3em]
\texttt{arr159@pitt.edu} \quad
\texttt{\{qiangzhang, zhuokai, lizhu\}@meta.com}
}

\begin{document}
\maketitle
\begin{abstract}

Large language models (LLMs) exhibit strong reasoning capabilities but typically require expensive post-training to reach high performance.
Recent test-time alignment methods offer a lightweight alternative, but have been explored mainly for preference alignment rather than reasoning.
To bridge this gap, we propose 
\textit{Token-level Adaptive Routing (\algname) }, which steers frozen LLMs toward structured reasoning entirely at inference time.
Specifically, we first train reward models on step-wise mathematical traces to capture fine-grained logical consistency signals, then introduce a learnable token-level router that automatically controls the guidance of the reward model to the base model. 
Extensive experiments show that \algname significantly improves reasoning performance by up to +22.4\% over base model and +8.4\% over existing token-level test-time alignment methods, while also boosting out-of-distribution clinical reasoning (MedXpertQA) and instruction following (AlpacaEval).
Furthermore, \algname also generalizes from small to large backbones without retraining, extending test-time alignment from preference optimization to robust, cross-domain reasoning.
%

%
\end{abstract}

\section{Introduction}

\begin{figure}
\centering
    \includegraphics[width=\columnwidth]{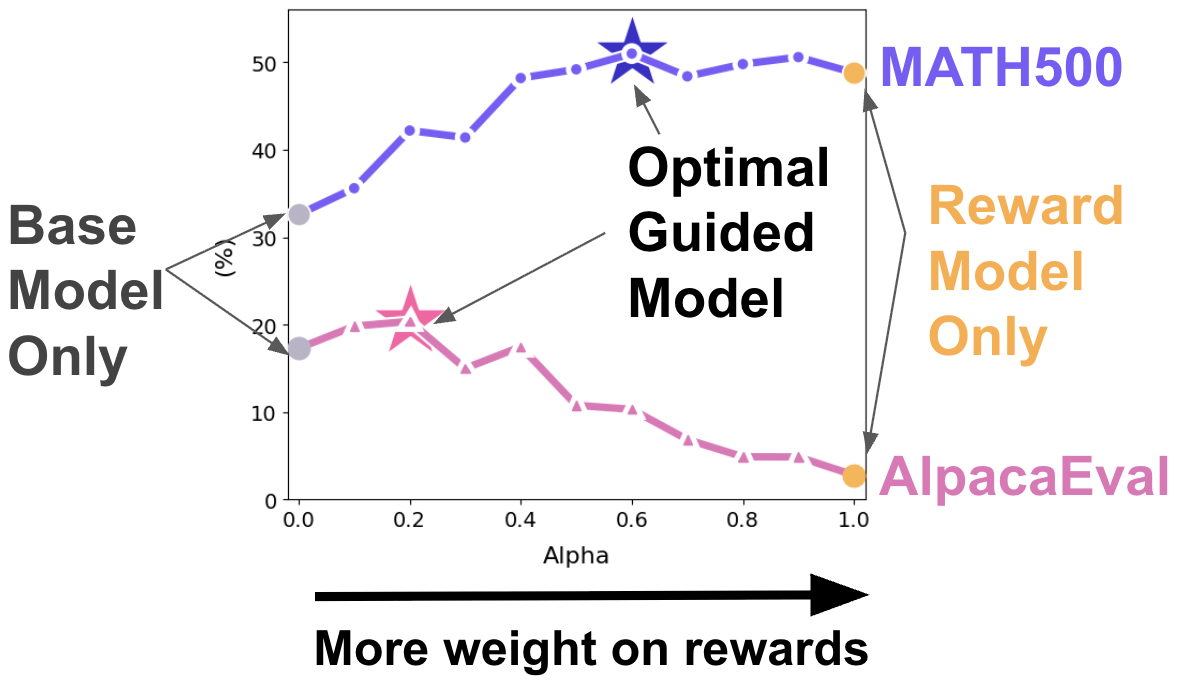}
  \caption{
  Performance on MATH500 (accuracy) and AlpacaEval (length-controlled win rate) for the state-of-the-art test-time alignment approach (GenARM) under different mixing coefficients $\alpha \in [0,1]$.
  An $\alpha=0$ corresponds to decoding solely from the base model, while $\alpha=1$ uses only the reward model.
  %
  }
  \label{fig:concept}
\end{figure}

Large Language Models (LLMs) have achieved impressive performance across many natural language tasks~\citep{openai2024learning, guo2025deepseek, team2025kimi, yang2025qwen3}.
On complex domains, such as mathematics, science, and clinical reasoning, it remains challenging to reliably solve logically demanding problems~\citep{mirzadeh2024gsm, wang2025survey, cui2025curie, wang2025medical,xiong2026token}.
Recent advances in LLMs post-training, especially reinforcement learning with verifiable reward (RLVR) approaches, such as group relative policy optimization (GRPO)~\citep{shao2024deepseekmath}, dynamic sampling policy optimization (DAPO)~\citep{yu2025dapo} and others~\citep{zheng2025group, yang2025let}, have substantially improved reasoning.
However, post-training approaches require costly model updates~\citep{casper2023open, hou2024does}, tend to be domain-specific~\citep{wu2025knowledge, qi2024quantifying}, and often degrade non-reasoning capabilities or disrupt previously learned user preferences~\citep{chen2024preference, xiao2025algorithmic}.
Moreover, retraining becomes increasingly impractical for larger LLMs and especially prohibitive when robust reasoning is needed across multiple, frequently changing domains.

Test-time alignment offers a lighter, 
and versatile alternative by steering a frozen LLM (the \textit{base} model) during decoding with a \textit{reward} model (usually a smaller LLM) that provides domain expertise or user preference signals complementary to the base model~\citep{pan2025survey, zhang2025survey}.
While this paradigm avoids costly retraining, existing approaches typically rely on fixed interpolation weights between the base and reward models~\citep{Xu2024GenARMRG}, requiring manual tuning and offering no mechanism to adapt guidance as decoding unfolds or as domains change.
%
%
In general-purpose deployments, where a model must handle diverse requests across tasks and domains, this rigidity becomes a significant limitation. Furthermore, as base models are scaled \cite{Xu2024GenARMRG}, the optimal balance between base and reward model guidance shifts, yet fixed interpolation weights offer no mechanism to accommodate this.

As shown in \figref{fig:concept}, the performance of existing token-level test-time alignment is  sensitive to hyperparameter choices, and the optimal hyperparameter varies substantially across domains and model families. 
For instance, fixing the interpolation weight at $\alpha=0.5$ as in GenARM~\citep{Xu2024GenARMRG} can even cause the guided model to underperform the base model on certain tasks.
Although test-time alignment is attractive for its flexibility, this lack of adaptive control over reward guidance limits robustness when transferring across tasks, domains, and model scales.

In this work, we propose \textbf{Token-level Adaptive Routing (\algname)}, that enables robust reasoning improvement without retraining the base model.
We first train the reward model on step-wise mathematical reasoning traces to capture fine-grained logical consistency signals.
To make this reward guidance effective across domains and model scales, we introduce a learnable token-level router that dynamically combines the base and reward model outputs at each decoding step, eliminating the need for manual hyperparameter tuning and improving stability when transferring across tasks, domains, and model families.

We evaluate \algname on both reasoning and non-reasoning benchmarks, including MATH500~\citep{lightman2023lets} for mathematical reasoning, MedXpertQA~\citep{Zuo2025MedXpertQABE} for clinical reasoning, and AlpacaEval~\citep{alpaca_eval} for instruction following. 
Our method consistently outperforms state-of-the-art test-time alignment methods, achieving up to +22.4\% accuracy over the base model and +8.4\% over GenARM~\citep{Xu2024GenARMRG} on MATH500~\citep{lightman2023lets}, while also delivering robust gains on out-of-distribution tasks such as clinical reasoning and instruction following.
Notably, the proposed router exhibits \textit{weak-to-strong generalization}: when trained on smaller models, it transfers effectively to much larger backbones (base models) without retraining, indicating that the learned token-level modulation is both scale- and architecture-agnostic.

To summarize, our contributions are threefold:
\begin{itemize}[leftmargin=*]
    \item[\circone] \textit{Token-level reasoning rewards:} we show that step-wise mathematical traces can train effective reward models for test-time reasoning guidance.
    %
    \item[\circtwo] \textit{Adaptive token-level router:} we propose a lightweight, learnable router that removes the need for manual interpolation tuning by dynamically blending base and reward logits.
    %
    \item[\circthree] \textit{Robust, transferable reasoning:} \algname consistently improves reasoning across domains and model scales without additional training, extending test-time alignment from preference optimization to general, cross-domain reasoning.
\end{itemize}

\section{Related Work}
\paragraph{Test-time alignment.} Expensive policy optimization methods have motivated a shift toward dynamic alignment approaches that operate during inference. Some TTA methods, such as Best-of-N sampling \cite{Gao2022ScalingLF}, rely on trajectory-level rewards and require multiple complete forward passes. Others apply trajectory-level reward models at each decoding step over full rollouts \cite{Chakraborty2024TransferQS, Huang2024DeALDA} or partial rollouts \cite{Khanov2024ARGSAA, Li2024CascadeRS}, making them prohibitively costly. In contrast, GenARM~\cite{Xu2024GenARMRG} learns to predict token-level rewards from preference data, eliminating the need for rollouts altogether. Concurrent work to ours, UniR \cite{Kim2025UniversalRA} also explores test-time alignment for reasoning, training a reward model with GRPO \cite{shao2024deepseekmath} alongside a frozen base model. Our approach is more similar to GenARM: we learn a mathematical reasoning reward model from preference data, independently of the base model. Beyond both GenARM and UniR, we further study how to achieve robust and adaptive test-time reasoning without relying on fixed reward and base model interpolation.

\paragraph{Post-training methods for reasoning.} Supervised finetuning (SFT) \cite{Guha2025OpenThoughtsDR, Li2024Common7L} has been used to enhance reasoning ability during post-training from datasets distilled from more advanced models \cite{DeepSeekAI2025DeepSeekR1IR, openai2024gpt4} or carefully curated datasets like \cite{Yue2023MAmmoTHBM, Ye2025LIMOLI}. Recently, reinforcement learning from verifiable rewards \cite{shao2024deepseekmath, Yu2025DAPOAO, Liu2025UnderstandingRT, Zheng2025GroupSP} have been used to significantly improve the reasoning ability of large language models. Our method seeks to improve the reasoning ability of LLMs as well, but does not require training the base or policy model.

\paragraph{Mixture of Experts.}
Mixture-of-Experts models have recently emerged as the state-of-the-art architecture for improving LLM capacity \cite{switch, deepseek-moe, yang2025qwen3}. 
In MoE, each expert specializes in a task domain \cite{btm,btx} where a router selects the most suitable experts for different input data. 
Recently, MoE has also been applied as adapters \cite{mixlora,hydralora,smore} on top of a frozen base model in parameter-efficient fine-tuning use cases. 
However, most existing works need to jointly train experts and router, where the router performs expert selection based on the model's hidden embeddings. 
Such designs make it less flexible if we were to replace experts in test time. 
In this work, we adapt the idea of MoE for test-time alignment, where we treat the base and reward models as experts and instantiate a router which is trained separately. 
We tailor the routing mechanism so that no re-training is needed when 
scaling up base models, leading to flexible and lightweight test-time alignment. 

\section{Method}

\subsection{Preliminaries}
Our work builds on GenARM~\citep{Xu2024GenARMRG}, which reformulates trajectory-level reward $r(x, y)$, a scalar score assigned to a full input-output pair, into token-level rewards produced by a reward model.
Formally, let $x \in \mathcal{X}$ denote a prompt (input sequence), and let $y = (y_{1}, \ldots, y_{|y|}) \in \mathcal{Y}$ denote the completion (LLM response) as a sequence of tokens.
GenARM models the reward as the log-likelihood of the trajectory under a reward-parameterized language model $\pi_{\text{reward}}$, i.e.
\begin{equation}
    r(x,y) \;=\; \sum_{t=1}^{|y|} \log \pi_{\text{reward}}(y_t \mid x, y_{<t}).\nonumber
    \label{eq:reward}
\end{equation}
where $y_{<t} = (y_1, \ldots, y_{t-1})$ denotes the prefix up to step $t-1$.
%
In GenARM, the reward model $\pi_{\text{reward}}$ is trained on human preference data~\citep{Ouyang2022TrainingLM} using a preference loss. 
At inference, GenARM combines the base model and the reward model at the token level. 
Specifically, the next-token distribution is given by a weighted sum of the base and reward model distributions:
\begin{align}
\pi_{\text{guided}}(y_t \mid x, y_{<t}) &= \pi_{\text{base}}(y_t \mid x, y_{<t}) \nonumber \\
&\quad + \alpha\,\pi_{\text{reward}}(y_t \mid x, y_{<t}),\nonumber
\label{eq:guided-decoding}
\end{align}
where $\alpha$ is a scalar controlling the influence of the reward model.
%
Next, we introduce our proposed method, \algname.
At a high-level, we first train a \textit{reasoning reward model}, and then learn a lightweight \textit{token-level router} that adaptively combines the logits from the base and reward models during decoding. 
We detail each component below.
%

\subsection{Reasoning Reward LLM}
\label{sec:math-reasoning-reward}
Unlike GenARM~\citep{Xu2024GenARMRG}, which learns rewards from preference pairs without explicitly modeling reasoning, we train a reasoning-aware reward model that directly targets stepwise logical correctness.
In practice, we use the Math-StepDPO-10K~\citep{Lai2024StepDPOSP} dataset, which captures fine-grained reasoning dynamics by constructing preference pairs between two step completions that share an identical correct reasoning \textit{prefix} but diverging at the next step, yielding one logically valid continuation $y_w$ and one incorrect continuation $y_l$.
Our reasoning reward model is thus optimized to prefer logically coherent, step-by-step reasoning over erroneous continuations by minimizing the standard preference loss~\cite{Ouyang2022TrainingLM, Bradley1952RankAO}:
%
%

\scalebox{0.8}{%
\begin{minipage}{\linewidth}
\centering
\begin{align}
l_{\text{pref}} =
    - \log \sigma \Big(\beta_r r(y_w,[x, \text{prefix}])-\nonumber
\beta_r r(y_l,[x,\text{prefix}])  \Big),
\end{align}
\vspace{0.01in}
\end{minipage}
}
%
\noindent where $[x, \text{prefix}]$ denotes the concatenation of the question $x$ and the shared correct reasoning prefix, $\beta_r$ is a temperature-like scaling factor, and $\sigma(\cdot)$ is the logistic sigmoid.
This objective encourages the reward model $\pi_{\text{reward}}$ to assign higher scores to steps that continue the reasoning correctly and lower scores to invalid ones.
%
%
%

\begin{figure*}[ht]
\centering
  \includegraphics[width=\linewidth]{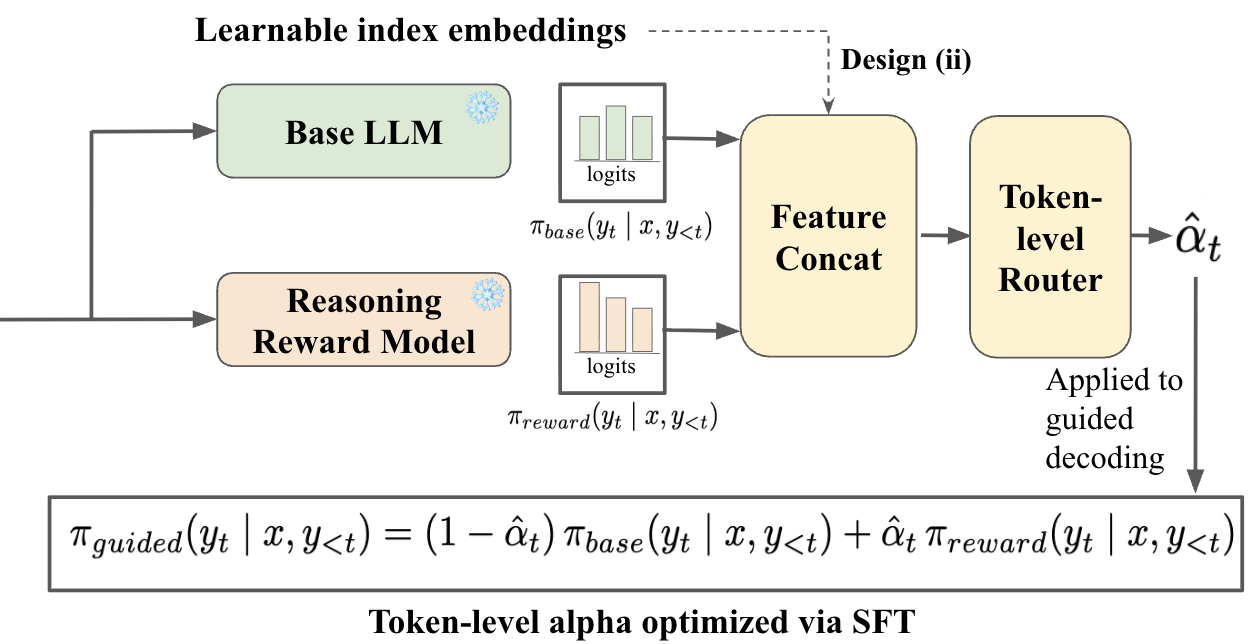}
  \caption{
    Learnable token-level router design.
    At each LLM decoding step $t$, the base and reward models produce logits $z^{\text{base}}_t$ and $z^{\text{reward}}_t$. 
    The logits are passed as input to \textit{Feature Concat}, which either (i) concatenates logits, or (ii) concatenates logits plus learnable token-index embeddings (as discussed in \secref{subsec: router}).
    The router consumes the concatenated feature and outputs a routing weight $\alpha_t \in (0,1)$.
    The guided distribution $(1-\alpha_t)\,z^{\text{base}}_t+\alpha_t\,z^{\text{reward}}_t$ is then used for sampling next token.
    This design makes the router portable across base model scales and families.
    %
  }
  \label{fig:learnable-router}
\end{figure*}

To reconcile step-level supervision with the token-level reward parameterization, we treat each step $y = (y_{1},\dots, y_{|y|})$ as a short trajectory and decomposes its scalar reward into per-token log-likelihood under the reward model.

\scalebox{0.9}{%
\begin{minipage}{\linewidth}
\centering
\begin{equation}
    r(y,[x,\text{prefix}])=\sum_{t=1}^{|y|}\log \pi_{\text{reward}}(y_t\mid [x,\text{prefix}],y_{<t}),\nonumber
\end{equation}
\vspace{0.01in}
\end{minipage}
}

\noindent where $y_{<t} = (y_1, \dots, y_{t-1})$.
This decomposition enables step-level preferences to supervise token-wise reward signals, preserving the fine-grained token-level formulation while aligning it with reasoning-driven correctness rather than flat, response-level preferences.
%

\subsection{Learnable Token-level Router}\label{subsec: router}
A fixed interpolation between base and reward model logits, as shown in \figref{fig:concept}, is fragile: one global mixing coefficient (i.e., $\alpha$) must simultaneously work across domains, model scales, and every decoding step.
To make the alignment adaptive, we introduce a lightweight token-level router that dynamically chooses how much to trust and utilize each model at every generation step.

Concretely, at decoding step $t$, the frozen base and reward models produce logits $z_t^{\text{base}}$ and $z_t^{\text{reward}}$.
A lightweight feed-forward router $g_\theta$ takes as input features derived from these two logits and outputs a scalar adaptive value $\alpha_t$.
We define this routing coefficient as $\alpha_t = \sigma(\alpha_t) \in (0, 1)$, which determines how much the guided decoding should follow the reward model versus the base model:
\begin{equation}
    z_{\text{guided}}(\cdot \mid x, y_{<t}) = (1 - \alpha_t)\,z_t^{\text{base}} + \alpha_t\,z_t^{\text{reward}},\nonumber
\end{equation}
Essentially, the proposed $\alpha_t$ adaptively adjusts the influence of the reward model token by token, in contrast to a fixed $\alpha$ that is expected to work universally across domains and model scales.

To prepare the router's input for predicting $\alpha_t$, we investigate two feature constructions that fundamentally differ in whether token position (index) information from the base and reward models is explicitly encoded.
%

\paragraph{Full-logits concatenation.}
In this design, we concatenate both logits from the base and reward model, before passing them through a small multi-layer perceptron (MLP), i.e.
\begin{equation}
    h_t^{\text{full}} = \big[z_t^{\text{base}} \,;\, z_t^{\text{reward}}\big] \in \mathbb{R}^{2V},
\end{equation}
where $V$ is the size of vocabulary.
This design is straightforward and utilizes the existing logit distributions of both models.
%

\paragraph{Top-$k$ logits with index embedding.}
Instead of operating purely in the raw logits space, we also consider explicitly encoding token index information by pairing each selected logit $z_{t, i}$ with a learnable index embedding $e_{t, i}$.
For each chosen token $z_{t,i}$ from the base model we form a feature vector that combines its logit value with its embedding; the same is done for tokens $j$ from the reward model.
Formally, we have:
\begin{align}
&u_{t,i}^{\text{base}} = \big[z^{\text{base}}_{t,i} \,;\, e_i\big]\in\mathbb{R}^{d+1},\nonumber \\
&u_{t,j}^{\text{reward}} = \big[z^{\text{reward}}_{t,j} \,;\, e_j\big]\in\mathbb{R}^{d+1}.\nonumber
\end{align}
Here $e_{i}=\mathbb{E}(i)$ comes from a $d$-dimensional learnable encoder $\mathbb{E}$, allowing the router to represent each token's identity rather than treating all tokens the same in the logit vector.
Since the position information is explicitly encoded, in practice we can restrict the inputs to only the top-$k$ tokens from each model.
This keeps the feature representation compact while preserving the most informative candidates for routing.

We then concatenate all index-augmented features from both models into a single vector:
\begin{equation}
    h_t^{\text{top-$k$}} = \big[
u_{t,1}^{\mathrm{base}},\dots,\,u_{t,k}^{\mathrm{base}} \;;\;
u_{t,1}^{\mathrm{reward}},\dots,\,u_{t,k}^{\mathrm{reward}}
\big]\nonumber
\end{equation}
This vector $h_t^{\text{top-$k$}} \in \mathbb{R}^{2K(d+1)}$ is passed through the same MLP as in the full-logits design to produce $\alpha_t$. 
Note that $k$ is meant to be very small, i.e., $|h_t^{\text{top-$k$}}| \ll |h_t^{\text{full}}|$.

\paragraph{Router design.}
In both cases, the resulting representation $h_t$ is passed through the same shallow MLP to predict the routing weight:
\begin{equation}
    \hat\alpha_t = \sigma\left(W_2\,\phi(W_1 h_t + b_1)+b_2\right),\nonumber
\end{equation}
where $\phi$ is the Tanh activation and $\sigma$ is the sigmoid function, constraining $\hat\alpha_t\in(0,1)$.

To promote confident routing behavior, we optionally add an entropy regularizer on $\hat\alpha_t$. 
This encourages the router to avoid indecisive values (e.g., $\hat\alpha_t \approx 0.5$) when the base and reward models diverge, thereby helping it to commit to the source it considers more reliable.
%

The overall training objective combines standard negative log-likelihood with the entropy penalty:
\begin{align}
\mathcal{L}_{\text{router}}
&= - \sum_{t} \log \pi_{\text{guided}}(y_t^\star \mid x, y_{<t}) \nonumber \\
&\quad + \lambda_{\text{entropy}} \sum_{t} H(\hat\alpha_t),\nonumber
\end{align}
where $y_t^\star$ is the gold target token and 
\begin{equation}
H(\hat\alpha_t) = -\hat\alpha_t \log \hat\alpha_t \nonumber - (1-\hat\alpha_t)\log(1-\hat\alpha_t)\nonumber
\end{equation}
is the Bernoulli entropy of the router's decision. 
The hyperparameter $\lambda_{\text{entropy}} \geq 0$ controls the strength of this confidence regularization. 

Note that \textit{no ground truth values of $\hat\alpha_t$ are required}. Instead, the router implicitly optimizes $\hat\alpha_t$ through $\mathcal{L}_{\text{router}}$: (1) the NLL term penalizes routing decisions that reduce the likelihood of gold tokens and (2) the entropy term encourages hard routing decisions over mixing uniformly between the two models. 

\paragraph{Final guided decoding.} 
With the learned router, the decoding distribution becomes:
\begin{align}
\pi_{\text{guided}}(y_t \mid x, y_{<t}) & =  (1-\hat\alpha_t)\,\pi_{\text{base}}(y_t \mid x, y_{<t}) \nonumber \\
&\quad + \hat\alpha_t\,\pi_{\text{reward}}(y_t \mid x, y_{<t}).\nonumber
\label{eq:router-guided}
\end{align}
This allows dynamic token-level modulation of reward guidance, improving reasoning ability while mitigating performance drop across domains.

\section{Experiment}
\begin{table*}[t]
\setlength{\tabcolsep}{18pt}
\resizebox{\textwidth}{!}{%
  \centering
  \begin{tabular}{lcccc}
    \hline
    \textbf{Method} & \textbf{MATH500} & \textbf{MedXpertQA} & \textbf{AlpacaEval} & \textbf{Average} \\
    \hline
    \multicolumn{5}{l}{\textit{Llama-3.1-8B Instruct Base \quad + \quad DeepSeek Distill Llama-8B Reward}} \\
    \hline
    UniR~\cite{Kim2025UniversalRA}    & 51.2 & -- & -- & -- \\
    Llama-3.1-8B Instruct (Base)  & 32.0 & 13.0 & 17.3 & 20.8 \\
    DeepSeek Llama-8B (Reward)    & 47.2 & 11.6  & 2.8 & 20.5 \\
    GenARM~\cite{Xu2024GenARMRG}* & 49.2  & 11.2 & 10.8 & 23.7 \\
    \rowcolor{lightgray} \algname (Ours)               & \textbf{54.4} & \textbf{13.2} & \textbf{20.8} & \textbf{29.5}  \\
    \hline
        \multicolumn{5}{l}{\textit{Qwen-2.5-3B Instruct Base \quad + \quad Qwen-2.5-0.5B Instuct Reward}} \\
    \hline
    UniR~\cite{Kim2025UniversalRA}     & 64.8 & -- & -- & -- \\
    Qwen-2.5 3B (Base)         & 60.8 & \textbf{11.8} & \textbf{33.6 }& \textbf{35.3} \\
    Qwen-2.5 0.5B (Reward)         & 
24.2 & 8.0  & 4.5 & 12.2 \\
    GenARM~\cite{Xu2024GenARMRG}*   & 56.4 & 9.8 & 18.9 & 28.4 \\
    \rowcolor{lightgray} \algname (Ours)                 & \textbf{64.8} & \underline{9.8}  & \underline{21.6} & \underline{32.0} \\

    \hline
    
  \end{tabular}
  }
  \caption{\label{tab:diverse-methods}
    Performance across reasoning (MATH500, MedXpertQA) and instruction-following (AlpacaEval) benchmarks. 
    Reward models are trained on Math-StepDPO-10K described in \secref{sec:math-reasoning-reward}. 
    Results for UniR are taken from its original paper,
    * indicates evaluated with our Math-StepDPO-10K trained reasoning reward model.
  }
\end{table*}

\subsection{Experimental Setup}

\paragraph{Benchmarks.}
We evaluate \algname on two reasoning domains: \texttt{MATH500}~\cite{Lightman2023LetsVS}, which is in-distribution with respect to the router training recipe, and \texttt{MedXpertQA}~\cite{Zuo2025MedXpertQABE}, which is out-of-distribution.
We also include \texttt{AlpacaEval}~\cite{alpaca_eval}, a general instruction-following benchmark for the multi-domain experiment.  
AlpacaEval mainly consists of knowledge-intensive question answering, but also includes simpler reasoning tasks in mathematics and coding.

\paragraph{Models.}
We experiment with two model families: \texttt{Llama-3.1}~\cite{meta2024llama31blog} and \texttt{Qwen-2.5}~\cite{qwen2025qwen25technicalreport}, using their instruct variants unless otherwise stated.
For the reward models, we use \texttt{DeepSeek-R1-Distill-Llama 8B}~\cite{DeepSeekAI2025DeepSeekR1IR} and \texttt{Qwen2.5-0.5B}. 
Reward models are trained on the step-wise preference reasoning dataset as discussed in \secref{sec:math-reasoning-reward}.
%

\paragraph{Implementation details.}
Following \citet{Xu2024GenARMRG} and \citet{Kim2025UniversalRA}, we train a separate reward model for each base model family.
In terms of router, we train the learnable router on examples from \texttt{Math-StepDPO-10K}~\cite{Lai2024StepDPOSP} and \texttt{UltraFeedback}~\cite{Cui2023UltraFeedbackBL}.
More training details and hyperparameters are illustrated in \appref{app:exp_details}.

During decoding, following \citet{Xu2024GenARMRG}, we use standard sampling with temperature $t=0.5$ across all models. 
We generate up to 512 tokens for AlpacaEval and MedXpertQA, and up to 2,048 tokens for MATH500.  
Prompts used in our experiments are reported in \appref{app:eval-prompt}.
%

\paragraph{Baselines.}
We compare \algname against 
(i) the \textit{base models}, 
(ii) the \textit{reward models}, 
and (iii) two state-of-the-art test-time alignment methods: \texttt{GenARM}~\cite{Xu2024GenARMRG} and \texttt{UniR}~\cite{Kim2025UniversalRA}.
For GenARM, we use the same reward model trained on MATH-StepDPO-10K~\cite{Lai2024StepDPOSP} as in our method ($\alpha=0.5$, equivalent to equal base and reward weighting).

\subsection{Results Across Diverse Domains}
\label{sec:exp-diverse-domain}

\tabref{tab:diverse-methods} reports results across MATH500, MedXpertQA, and AlpacaEval. 
For the Llama-3.1 family, our method improves over both the base and reward models individually, and also outperforms GenARM. 
The largest gains appear on MATH500, where the reasoning reward model is extremely helpful with mathematical reasoning, and we also observe improvements on the out-of-distribution MedXpertQA domain.

For the Qwen-2.5 family, the base model is already very strong, especially on AlpacaEval. 
While our method does not outperform the Qwen-2.5 base model on AlpacaEval and MedXpertQA, it consistently exceeds GenARM across domains, showing that token-level routing provides more effective reward guidance than static interpolation. 
Importantly, our reward models are trained using a preference loss on \emph{step-wise mathematical preference data}, which is considerably simpler than the RL objective used in GRPO for UniR~\cite{Kim2025UniversalRA}.
Despite the relatively low standalone performance of the reward model, this does not imply a lack of utility: the reward model may be overfitting to signals of step-wise mathematical reasoning, which could remain highly beneficial when used to steer the base model. 
However, we are only able to demonstrate utility when this weak reward model is combined with token-level routing. 
Overall, we are able to show that effective \emph{token-level reward models for math} can be constructed directly from math step-wise reasoning traces preferences.
%

We also compare \algname against majority voting (N=8) and show that \algname achieves higher accuracy on MATH500 with approximately 4x less compute. 
The results are detailed in \appref{app:majority_voting}.
%

Interestingly, GenARM notably improved the performance of Llama-3.1-8B Instruct (Base) but failed to yield gains on Qwen-2.5 3B (Base) when evaluated on MATH500. 
We hypothesize that GenARM and other similar test-time alignment methods may be ineffective when the reward model performs substantially worse than the base model. 
In contrast, our proposed method adaptively controls the reward weight, which can significantly mitigate this limitation.


\subsection{Weak-to-Strong Generalization}
\label{sec:exp-weak-strong}
\begin{figure}[h]
  \includegraphics[width=\columnwidth]{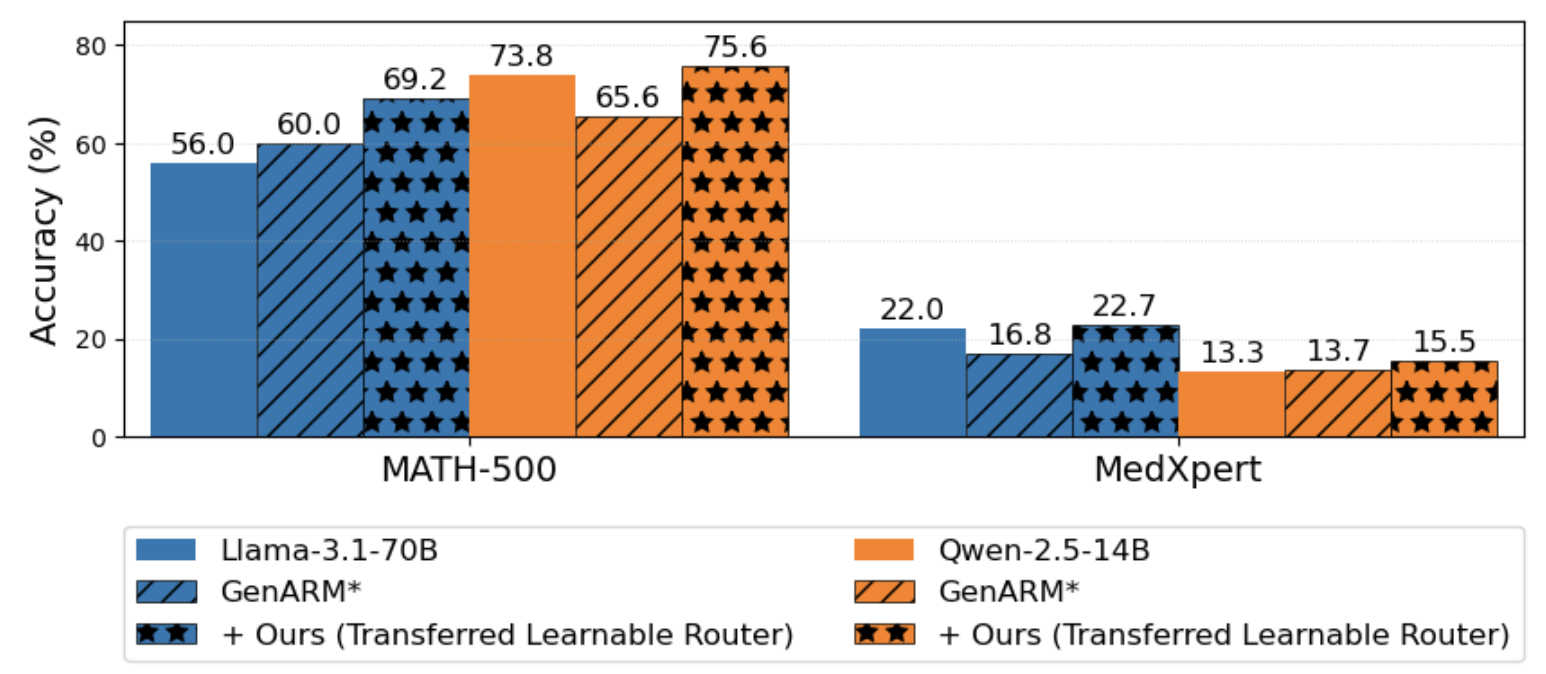}
  \caption{
    Weak-to-strong generalization of learned router on reasoning. Learned router and reasoning reward model are not retrained for this scale.
  }
  \label{fig:weak2strong}
\end{figure}

Using the router trained in \secref{sec:exp-diverse-domain}, we evaluate its transferability by scaling to larger base models \textit{without} any re-training. 
Specifically, we pair the learned router with Llama-3.1-70B and Qwen-2.5-14B backbones. 
As shown in \figref{fig:weak2strong}, our approach consistently improves over both the base and GenARM. 
On MATH500, the transferred router achieves substantial gains, while on MedXpertQA, improvements are smaller but still positive.

This setting highlights the weak-to-strong generalization property of both our router and reward model: routers and reward models trained on relatively small backbones can effectively steer much larger frozen LLMs. 
Importantly, despite being trained on limited step-wise mathematical preference data, the router provides transferable benefits even in out-of-domain reasoning tasks such as MedXpertQA.

\section{Analysis and Ablation Studies}
\label{sec:ablations}

\subsection{Understanding Router Behavior}
\begin{table}[t]
\small
  \centering
  \begin{tabular}{ll}
    \hline
    \textbf{High $\alpha$ Tokens} & \textbf{Low $\alpha$ Tokens} \\
    \hline
    \texttt{rightarrow}        & \texttt{period} \\
    \texttt{Exp}         & \texttt{-coordinate} \\
    \texttt{Step}        & \texttt{profit} \\
    \texttt{cases}   & \texttt{factors} \\
    \texttt{critical}          & \texttt{shaded} \\
    \texttt{Part} & \texttt{squares} \\
    \texttt{Find} & \texttt{students} \\
    \texttt{understand}           & \texttt{time} \\
    \texttt{begin}         & \texttt{digits} \\
    \texttt{Setting}        & \texttt{bag} \\
    \texttt{equiv} & \texttt{houses} \\
    \texttt{Evaluate} & \texttt{digit} \\
    \texttt{analyzing}         & \texttt{What} \\
    \texttt{opens}          & \texttt{mean} \\
    \texttt{back}      & \texttt{pets} \\
    \texttt{determine} & \texttt{sums} \\
    \texttt{compute}           & \texttt{bles} \\
    \texttt{end}         & \texttt{region} \\
    \texttt{follow} & \texttt{already} \\
    \hline
  \end{tabular}
  \caption{Tokens in top 0.1\% and bottom 0.1\% of $\alpha$ from generated responses to MATH500 responses; filtered out tokens with less than 50 occurrences and shorter than 2 characters (difficult to interpret). 
  Tokens on the left show strong reward model influence, reflecting mathematical operators, 
  formatting, and reasoning scaffolds (e.g. ``cases'', ``Step''). Tokens on the right are dominated by the base model, 
  largely consisting of vocabulary from the problem context. 
  Base Model = Qwen2.5-3B; Reward Model = Qwen2.5-0.5B.}
  \label{tab:alpha-top}
\end{table}

We further investigate which tokens are more influenced by the reward model by analyzing the learned $\hat\alpha_t$ values. 
\tabref{tab:alpha-top} shows tokens from the top 0.1\% and bottom 0.1\% of average $\hat\alpha_t$ on MATH500. 
High-$\alpha$ tokens include mathematical operators, formatting, and scaffolding symbols (e.g., \texttt{``cases''}, \texttt{``Step''}), where the reward model contributes most strongly. 
In contrast, low-$\alpha$ tokens consist mainly of problem context tokens from the base model, such as common words (\texttt{``period''}, \texttt{``students''}, \texttt{``already''}). 
%
%
This analysis shows that the router allocates reward-model guidance to tokens that shape the underlying reasoning process, while relying on the base model to preserve fluent language and accurately convey problem-specific details.
\begin{table}[t]
\centering
\small
\begin{tabular}{l c}
\toprule
\textbf{Router Input Feature} & \textbf{MATH500 Accuracy (\%)} \\
\midrule
Reward hidden state only & 51.2 \\
Base + reward hidden states & 49.6  \\
Reward logits only & 52.6 \\
Base + reward logits & \textbf{54.4} \\
\bottomrule
\end{tabular}
\caption{Ablation on router input features for Llama-3.1-8B Instruct with DeepSeek Distill Llama-8B reward. Results are reported on MATH500. 
See \appref{app:router_inputs} for results on all three benchmarks.}
\label{tab:router-features}
\end{table}

\begin{table}[t!]
\centering
\small
\begin{tabular}{l l c c}
\toprule
\textbf{Model} & \textbf{Variant} & \textbf{MATH} & \textbf{MedX} \\
\midrule
Qwen & $K$ = 100 & 60.2 & 7.7 \\
Qwen & $K$ = 1,000 & 60.2 & 7.3 \\
Qwen & $K$ = 10,000 & 61.8 & 7.4 \\
Qwen & $K$ = 100 + $\lambda_{\text{entropy}}=0.5$ & 59.8 & 7.3 \\
Qwen & $K$ = 1,000 + $\lambda_{\text{entropy}}=0.5$ & \textbf{64.8} & \textbf{9.8} \\
\midrule
Llama & $K$ = 1,000 + $\lambda_{\text{entropy}}=0.5$ & 50.6 & \textbf{14.5} \\
Llama & Full logits & \textbf{54.4} & \underline{13.2} \\
\bottomrule
\end{tabular}
\caption{Ablation on Top-$K$, entropy regularization, and full-logit router designs. Accuracy (\%) is reported on MATH500 (MATH) and MedXpertQA (MedX). }
\label{tab:topk-ablation}
\end{table}

\subsection{Qualitative Analysis}
\begin{table*}[h]
\small
  \centering
  \begin{tabular}{p{1.4cm}p{6cm}p{3.5cm}p{3.5cm}} \hline
      \multicolumn{4}{p{\linewidth}}{\textbf{Question:} A worker receives an annual wage of $\$20{,}000$, which he always deposits into a savings account at the end of the year. By the end of the third year (when he makes the third deposit), he wants to have at least $\$66,200$ in the account to finance the purchase of a house. What is the minimal compound interest rate that the savings account must provide? Express your answer as a percentage, but do not include the percent sign. 
      } \\
      \multicolumn{4}{p{\linewidth}}{\textbf{Ground Truth:} 10} \\
    \hline
     & \textbf{Base Model} & \textbf{Reward Model} & \textbf{Ours} \\
    \hline
    \textbf{Output (abridged)} &
    Let r denote the interest rate. The worker will have a total of $20,000(1+r)^3$ in the account by the end of the third year.
    This is at least $66,200$, so we have the inequality:
    $
    \begin{aligned} 
        \\20{,}000(1+r)^3 &\ge 66{,}200 \\
        (1+r)^3 &\ge 3.31 \\ 
        1+r &\ge \sqrt[3]{3.31} \\ 
        r &\ge \sqrt[3]{3.31} - 1 \\ 
        r &\ge 0.322 \\\\
    \end{aligned}
    $
    
    Thus, the smallest possible value of $r$ is 0.322, or 32.2.
    Final Answer: $\boxed{32.2}$. 
    I hope it is correct.
    &Let the annual interest rate be $r$. 
    The worker makes three deposits of \$20,000 at the end of each year.
    After the first year: $20,000 (1 + r)$.  
    After the second year: $20,000 (1 + r)^2 + 20,000 (1 + r)$.  
    After the third year: $20,000 (1 + r)^3 + 20,000 (1 + r) + 20,000 (1 + r)$.  
    Wait, this seems incorrect$\dots$ 
    Wait, no$\dots$ The correct amount$\dots$  
    \textbf{[... stuck in confusion]}
    &Step 1: The worker deposits \$20,000 each year, wants at least \$66,200 after three years.

    Step 2: Total after 3 years is $20,000(1+r)^3 + 20,000(1+r)^2 + 20,000(1+r)$.  

    Step 3: Divide by 20,000: $(1+r)^3+(1+r)^2+(1+r)\geq 3.31$.

    Step 4: Let $x=1+r$, then $x^3+x^2+x \geq 3.31$.  

    Step 5: Solving gives $x \geq 1.1$, so $r \geq 0.1$.  

    Final Answer: $\boxed{10}$ \\\hline
  \end{tabular}
  \caption{Qualitative example of base (Llama-3.1 8B Instruct), reward and our model outputs (abridged to fit) for a MATH500 question. The base model oversimplifies, while the reward model understands the right plan but gets confused. Ours solves correctly by engaging in reasoning steps.}
  \label{tab:model-comparison}
\end{table*}

To better understand model behavior, \tabref{tab:model-comparison} shows responses from the base model, reward model, and our method to a problem from MATH500. 
The base model over-simplifies and arrives at an incorrect numeric answer, while the reward model identifies the correct reasoning path but becomes stuck mid-derivation. 
Our method succeeds by explicitly decomposing the reasoning into interpretable steps, ultimately producing the correct final answer. 
This illustrates the benefit of dynamically leveraging both base and reward signals at the token level.
We present more examples in \appref{appx:addiotion-qual-ex}.

\subsection{Ablations on Token-level Router}
\label{subsec:ablations_on_token-level_router}

We first validate the necessity of token-level routing granularity.
As shown in \appref{app:token_vs_prompt}, a prompt-level router that predicts a single $\alpha$ for the entire sequence significantly under-performs token-level routing on MATH500 (33.2\% vs. 49.6\%), confirming that fine-grained, per-token control is essential for structured reasoning.
Having established token-level routing, we next ablate the choice of input features for the router
%
and effect of restricting router inputs to the Top-k logits.

\paragraph{Router input feature choice.} 
On Llama-3.1-8B Instruct and DeepSeek Distill Llama-8B reward, we experimented with several input choices for the router: 
(1) reward hidden state only, 
(2) base and reward hidden states, 
and (3) reward logits. 
Since hidden states encode token-level context, it may be natural to consider them as router inputs. 
As shown in \tabref{tab:router-features}, we found that using the reward hidden state alone outperforms adding the base hidden state on MATH500 (51.2\% vs 49.6\%), while on out-of-domain tasks adding the base hidden state improves performance (18.7\% vs 15.7\% on AlpacaEval and 13.7\% vs 12.6\% on MedXpertQA). 
However,  base and reward hidden-state-based routers cannot reliably generalize to stronger base models, as the base hidden-state distribution shifts with model scale. 
This motivated our use of logits, which are scale- and domain-agnostic since they reflect model confidence over the predicted token distribution. 
We also found using base and reward logits yields the strongest results on MATH500 (54.4\%).

\paragraph{Top-$k$ ablation.}  
We next study the effect of restricting the router inputs to the Top-$k$ logits. 
The motivation for this design is to reduce noise from the full vocabulary distribution and focus the router on the most confident token candidates. 
As shown in \tabref{tab:topk-ablation}, on Qwen-2.5, we find that using $K=1000$ performs well, and adding entropy regularization further improves MATH500 accuracy to 64.8\%. 
The entropy penalty encourages the router to make more decisive choices between the base and reward models, which is particularly effective when the reward model provides complementary signal.  

On Llama-3.1, however, we observe that the entropy regularization is suboptimal on MATH500. 
Note in \tabref{tab:diverse-methods}, the reasoning reward model performs relatively closer to the base model on MATH500 than on AlpacaEval. 
This could suggest that on most of the soft router training dataset, the base model has a balanced performance, and this could drive the router to predict consistently smaller alpha values, since the base model is more consistent in it's performance. 
Possibly, the reasoning reward model is then underused in this configuration.
%
This effect could suggest that the benefit of entropy regularization is sensitive to the relative strengths of the base and reward models.
%

\subsection{Inference Cost}
We compare the inference-time efficiency of \algname against GenARM in \tabref{tab:throughput}, measured on a single node with 8×H100 GPUs. 
\algname introduces a lightweight router module to dynamically adjust $\alpha$, and we find that throughput is comparable to GenARM when the Top-K logits (Design ii in \secref{subsec: router}) is used.
Expanding to full logits (Design i in \secref{subsec: router}) reduces throughput, particularly for large-vocabulary models such as \texttt{Llama3.1-8B}; however, vocabulary sizes appear to have stabilized in recent model releases, suggesting this overhead will not grow with newer models.
We report both tokens-per-second (TPS) and queries-per-second (QPS). The latter does not penalize concise generations unlike TPS, making it a more complete measure of end-to-end throughput. In terms of QPS, the overhead is minimal in both settings and even lower than GenARM in the Top-K setting due to more concise outputs.
%
%
More details are discussed in \appref{app:router_complexity}.
%
\begin{table}[t]
\centering
\small
\setlength{\tabcolsep}{2pt}
\begin{tabular}{llccc}
\toprule
\textbf{Model} & \textbf{Method} & \textbf{TPS} & \textbf{GPU Mem. (\%)} & \textbf{QPS} \\
\midrule
\multirow{2}{*}{Qwen2.5-3B}  & GenARM              & 85.6 & 4.29  & 0.0651 \\
                              & \algname (Top-K) & 85.9 & 5.22  & 0.0448 \\
\midrule
\multirow{2}{*}{Llama3.1-8B} & GenARM              & 80.7 & 6.93  & 0.0141 \\
                              & \algname (Full)  & 32.5 & 10.45 & 0.0161 \\
\bottomrule
\end{tabular}
\caption{Throughput analysis on MATH500. TPS = tokens per second; QPS = queries per second. Note that these results reflect unoptimized implementations;  techniques such as speculative decoding and fully sharded data-parallel inference have not been applied.}
\label{tab:throughput}
\end{table}
\section{Conclusion}

In this paper, we introduced \algname, a test-time alignment framework that improves LLM reasoning by adaptively routing between a frozen base model and a reward model at the token level. 
Across mathematical reasoning, clinical reasoning, and instruction-following benchmarks, \algname consistently outperforms fixed-weight decoding baselines while preserving the flexibility and low training cost of inference-time alignment. 
Our results show that fine-grained reward signals, even when trained from step-wise mathematical preference data, can generalize beyond their source domain when applied through adaptive routing rather than static interpolation. 
We further find that the learned routing policy transfers to larger backbones without retraining, suggesting that token-level logit routing provides a scalable and portable interface for test-time reasoning control. 
Overall, \algname highlights that lightweight adaptive routing can be a practical path toward stronger, more robust reasoning in frozen LLMs without expensive post-training.
%



\clearpage
\bibliography{custom}

@misc{openai2024learning,
  title = {Learning to Reason with LLMs},
  author = {{OpenAI}},
  year = {2024},
  howpublished = {\url{https://openai.com/index/learning-to-reason-with-llms/}},
  note = {Accessed: 2025-05-01}
}

@article{guo2025deepseek,
  title={Deepseek-r1: Incentivizing reasoning capability in llms via reinforcement learning},
  author={Guo, Daya and Yang, Dejian and Zhang, Haowei and Song, Junxiao and Zhang, Ruoyu and Xu, Runxin and Zhu, Qihao and Ma, Shirong and Wang, Peiyi and Bi, Xiao and others},
  journal={arXiv preprint arXiv:2501.12948},
  year={2025}
}

@article{team2025kimi,
  title={Kimi k1. 5: Scaling reinforcement learning with llms},
  author={Team, Kimi and Du, Angang and Gao, Bofei and Xing, Bowei and Jiang, Changjiu and Chen, Cheng and Li, Cheng and Xiao, Chenjun and Du, Chenzhuang and Liao, Chonghua and others},
  journal={arXiv preprint arXiv:2501.12599},
  year={2025}
}

@article{yang2025qwen3,
  title={Qwen3 technical report},
  author={Yang, An and Li, Anfeng and Yang, Baosong and Zhang, Beichen and Hui, Binyuan and Zheng, Bo and Yu, Bowen and Gao, Chang and Huang, Chengen and Lv, Chenxu and others},
  journal={arXiv preprint arXiv:2505.09388},
  year={2025}
}

@article{mirzadeh2024gsm,
  title={Gsm-symbolic: Understanding the limitations of mathematical reasoning in large language models},
  author={Mirzadeh, Iman and Alizadeh, Keivan and Shahrokhi, Hooman and Tuzel, Oncel and Bengio, Samy and Farajtabar, Mehrdad},
  journal={arXiv preprint arXiv:2410.05229},
  year={2024}
}

@article{wang2025medical,
  title={Medical Reasoning in the Era of LLMs: A Systematic Review of Enhancement Techniques and Applications},
  author={Wang, Wenxuan and Ma, Zizhan and Ding, Meidan and Zheng, Shiyi and Liu, Shengyuan and Liu, Jie and Ji, Jiaming and Chen, Wenting and Li, Xiang and Shen, Linlin and others},
  journal={arXiv preprint arXiv:2508.00669},
  year={2025}
}

@article{wang2025survey,
  title={A Survey on Large Language Models for Mathematical Reasoning},
  author={Wang, Peng-Yuan and Liu, Tian-Shuo and Wang, Chenyang and Wang, Yi-Di and Yan, Shu and Jia, Cheng-Xing and Liu, Xu-Hui and Chen, Xin-Wei and Xu, Jia-Cheng and Li, Ziniu and others},
  journal={arXiv preprint arXiv:2506.08446},
  year={2025}
}

@article{shao2024deepseekmath,
  title={Deepseekmath: Pushing the limits of mathematical reasoning in open language models},
  author={Shao, Zhihong and Wang, Peiyi and Zhu, Qihao and Xu, Runxin and Song, Junxiao and Bi, Xiao and Zhang, Haowei and Zhang, Mingchuan and Li, YK and Wu, Y and others},
  journal={arXiv preprint arXiv:2402.03300},
  year={2024}
}

@article{wu2025knowledge,
  title={Knowledge or Reasoning? A Close Look at How LLMs Think Across Domains},
  author={Wu, Juncheng and Liu, Sheng and Tu, Haoqin and Yu, Hang and Huang, Xiaoke and Zou, James and Xie, Cihang and Zhou, Yuyin},
  journal={arXiv preprint arXiv:2506.02126},
  year={2025}
}

@article{lightman2023lets,
      title={Let's Verify Step by Step}, 
      author={Lightman, Hunter and Kosaraju, Vineet and Burda, Yura and Edwards, Harri and Baker, Bowen and Lee, Teddy and Leike, Jan and Schulman, John and Sutskever, Ilya and Cobbe, Karl},
      journal={arXiv preprint arXiv:2305.20050},
      year={2023}
}

@article{Xu2024GenARMRG, 
  title={GenARM: Reward guided generation with autoregressive reward model for test-time alignment},
  author={Xu, Yuancheng and Sehwag, Udari Madhushani and Koppel, Alec and Zhu, Sicheng and An, Bang and Huang, Furong and Ganesh, Sumitra},
  journal={arXiv preprint arXiv:2410.08193},
  year={2024}
}

@article{zhang2025survey,
  title={A Survey on Test-Time Scaling in Large Language Models: What, How, Where, and How Well?},
  author={Zhang, Qiyuan and Lyu, Fuyuan and Sun, Zexu and Wang, Lei and Zhang, Weixu and Hua, Wenyue and Wu, Haolun and Guo, Zhihan and Wang, Yufei and Muennighoff, Niklas and others},
  journal={arXiv preprint arXiv:2503.24235},
  year={2025}
}

@article{pan2025survey,
  title={A Survey on Training-free Alignment of Large Language Models},
  author={Pan, Birong and Li, Yongqi and Zhang, Weiyu and Lu, Wenpeng and Xu, Mayi and Zhou, Shen and Zhu, Yuanyuan and Zhong, Ming and Qian, Tieyun},
  journal={arXiv preprint arXiv:2508.09016},
  year={2025}
}

@article{chen2024preference,
  title={Preference learning algorithms do not learn preference rankings},
  author={Chen, Angelica and Malladi, Sadhika and Zhang, Lily H and Chen, Xinyi and Zhang, Qiuyi and Ranganath, Rajesh and Cho, Kyunghyun},
  journal={Advances in Neural Information Processing Systems},
  volume={37},
  pages={101928--101968},
  year={2024}
}

@article{xiao2025algorithmic,
  title={On the algorithmic bias of aligning large language models with rlhf: Preference collapse and matching regularization},
  author={Xiao, Jiancong and Li, Ziniu and Xie, Xingyu and Getzen, Emily and Fang, Cong and Long, Qi and Su, Weijie J},
  journal={Journal of the American Statistical Association},
  number={just-accepted},
  pages={1--21},
  year={2025},
  publisher={Taylor \& Francis}
}

@article{qi2024quantifying,
  title={Quantifying generalization complexity for large language models},
  author={Qi, Zhenting and Luo, Hongyin and Huang, Xuliang and Zhao, Zhuokai and Jiang, Yibo and Fan, Xiangjun and Lakkaraju, Himabindu and Glass, James},
  journal={arXiv preprint arXiv:2410.01769},
  year={2024}
}

@article{hou2024does,
  title={Does RLHF Scale? Exploring the Impacts From Data, Model, and Method},
  author={Hou, Zhenyu and Du, Pengfan and Niu, Yilin and Du, Zhengxiao and Zeng, Aohan and Liu, Xiao and Huang, Minlie and Wang, Hongning and Tang, Jie and Dong, Yuxiao},
  journal={arXiv preprint arXiv:2412.06000},
  year={2024}
}

@article{casper2023open,
  title={Open problems and fundamental limitations of reinforcement learning from human feedback},
  author={Casper, Stephen and Davies, Xander and Shi, Claudia and Gilbert, Thomas Krendl and Scheurer, J{\'e}r{\'e}my and Rando, Javier and Freedman, Rachel and Korbak, Tomasz and Lindner, David and Freire, Pedro and others},
  journal={arXiv preprint arXiv:2307.15217},
  year={2023}
}

@article{yu2025dapo,
  title={Dapo: An open-source llm reinforcement learning system at scale},
  author={Yu, Qiying and Zhang, Zheng and Zhu, Ruofei and Yuan, Yufeng and Zuo, Xiaochen and Yue, Yu and Dai, Weinan and Fan, Tiantian and Liu, Gaohong and Liu, Lingjun and others},
  journal={arXiv preprint arXiv:2503.14476},
  year={2025}
}

@article{zheng2025group,
  title={Group sequence policy optimization},
  author={Zheng, Chujie and Liu, Shixuan and Li, Mingze and Chen, Xiong-Hui and Yu, Bowen and Gao, Chang and Dang, Kai and Liu, Yuqiong and Men, Rui and Yang, An and others},
  journal={arXiv preprint arXiv:2507.18071},
  year={2025}
}

@article{cui2025curie,
  title={CURIE: Evaluating LLMs On Multitask Scientific Long Context Understanding and Reasoning},
  author={Cui, Hao and Shamsi, Zahra and Cheon, Gowoon and Ma, Xuejian and Li, Shutong and Tikhanovskaya, Maria and Norgaard, Peter and Mudur, Nayantara and Plomecka, Martyna and Raccuglia, Paul and others},
  journal={arXiv preprint arXiv:2503.13517},
  year={2025}
}

@article{Ouyang2022TrainingLM,
  title={Training language models to follow instructions with human feedback},
  author={Long Ouyang and Jeff Wu and Xu Jiang and Diogo Almeida and Carroll L. Wainwright and Pamela Mishkin and Chong Zhang and Sandhini Agarwal and Katarina Slama and Alex Ray and John Schulman and Jacob Hilton and Fraser Kelton and Luke E. Miller and Maddie Simens and Amanda Askell and Peter Welinder and Paul Francis Christiano and Jan Leike and Ryan J. Lowe},
  journal={ArXiv},
  year={2022},
  volume={abs/2203.02155},
  url={https://api.semanticscholar.org/CorpusID:246426909}
}

@article{Lai2024StepDPOSP,
  title={Step-DPO: Step-wise Preference Optimization for Long-chain Reasoning of LLMs},
  author={Xin Lai and Zhuotao Tian and Yukang Chen and Senqiao Yang and Xiangru Peng and Jiaya Jia},
  journal={ArXiv},
  year={2024},
  volume={abs/2406.18629},
  url={https://api.semanticscholar.org/CorpusID:270764693}
}

@article{Zuo2025MedXpertQABE,
  title={MedXpertQA: Benchmarking Expert-Level Medical Reasoning and Understanding},
  author={Yuxin Zuo and Shang Qu and Yifei Li and Zhangren Chen and Xuekai Zhu and Ermo Hua and Kaiyan Zhang and Ning Ding and Bowen Zhou},
  journal={ArXiv},
  year={2025},
  volume={abs/2501.18362},
  url={https://api.semanticscholar.org/CorpusID:275993625}
}

@article{Lightman2023LetsVS,
  title={Let's Verify Step by Step},
  author={Hunter Lightman and Vineet Kosaraju and Yura Burda and Harrison Edwards and Bowen Baker and Teddy Lee and Jan Leike and John Schulman and Ilya Sutskever and Karl Cobbe},
  journal={ArXiv},
  year={2023},
  volume={abs/2305.20050},
  url={https://api.semanticscholar.org/CorpusID:258987659}
}

@misc{alpaca_eval,
  author = {Xuechen Li and Tianyi Zhang and Yann Dubois and Rohan Taori and Ishaan Gulrajani and Carlos Guestrin and Percy Liang and Tatsunori B. Hashimoto },
  title = {AlpacaEval: An Automatic Evaluator of Instruction-following Models},
  year = {2023},
  month = {5},
  publisher = {GitHub},
  journal = {GitHub repository},
  howpublished = {\url{https://github.com/tatsu-lab/alpaca_eval}}
}

@misc{meta2024llama31blog,
  author       = {{Llama Team}},
  title        = {Introducing Llama 3.1: Our Most Capable Models to Date},
  howpublished = {\url{https://ai.meta.com/blog/meta-llama-3-1/}},
  note         = {Meta AI Blog},
  year         = {2024}
}

@misc{qwen2025qwen25technicalreport,
      title={Qwen2.5 Technical Report}, 
      author={Qwen and : and An Yang and Baosong Yang and Beichen Zhang and Binyuan Hui and Bo Zheng and Bowen Yu and Chengyuan Li and Dayiheng Liu and Fei Huang and Haoran Wei and Huan Lin and Jian Yang and Jianhong Tu and Jianwei Zhang and Jianxin Yang and Jiaxi Yang and Jingren Zhou and Junyang Lin and Kai Dang and Keming Lu and Keqin Bao and Kexin Yang and Le Yu and Mei Li and Mingfeng Xue and Pei Zhang and Qin Zhu and Rui Men and Runji Lin and Tianhao Li and Tianyi Tang and Tingyu Xia and Xingzhang Ren and Xuancheng Ren and Yang Fan and Yang Su and Yichang Zhang and Yu Wan and Yuqiong Liu and Zeyu Cui and Zhenru Zhang and Zihan Qiu},
      year={2025},
      eprint={2412.15115},
      archivePrefix={arXiv},
      primaryClass={cs.CL},
      url={https://arxiv.org/abs/2412.15115}, 
}

@article{DeepSeekAI2025DeepSeekR1IR,
  title={DeepSeek-R1: Incentivizing Reasoning Capability in LLMs via Reinforcement Learning},
  author={DeepSeek-AI and Daya Guo and Dejian Yang and Haowei Zhang and Jun-Mei Song and Ruoyu Zhang and Runxin Xu and Qihao Zhu and Shirong Ma and Peiyi Wang and Xiaoling Bi and Xiaokang Zhang and Xingkai Yu and Yu Wu and Z. F. Wu and Zhibin Gou and Zhihong Shao and Zhuoshu Li and Ziyi Gao and Aixin Liu and Bing Xue and Bing-Li Wang and Bochao Wu and Bei Feng and Chengda Lu and Chenggang Zhao and Chengqi Deng and Chenyu Zhang and Chong Ruan and Damai Dai and Deli Chen and Dong-Li Ji and Erhang Li and Fangyun Lin and Fucong Dai and Fuli Luo and Guangbo Hao and Guanting Chen and Guowei Li and H. Zhang and Han Bao and Hanwei Xu and Haocheng Wang and Honghui Ding and Huajian Xin and Huazuo Gao and Hui Qu and Hui Li and Jianzhong Guo and Jiashi Li and Jiawei Wang and Jingchang Chen and Jingyang Yuan and Junjie Qiu and Junlong Li and Jiong Cai and Jiaqi Ni and Jian Liang and Jin Chen and Kai Dong and Kai Hu and Kaige Gao and Kang Guan and Kexin Huang and Kuai Yu and Lean Wang and Lecong Zhang and Liang Zhao and Litong Wang and Liyue Zhang and Lei Xu and Leyi Xia and Mingchuan Zhang and Minghua Zhang and M. Tang and Meng Li and Miaojun Wang and Mingming Li and Ning Tian and Panpan Huang and Peng Zhang and Qiancheng Wang and Qinyu Chen and Qiushi Du and Ruiqi Ge and Ruisong Zhang and Ruizhe Pan and Runji Wang and R. J. Chen and Ruiqi Jin and Ruyi Chen and Shanghao Lu and Shangyan Zhou and Shanhuang Chen and Shengfeng Ye and Shiyu Wang and Shuiping Yu and Shunfeng Zhou and Shuting Pan and S. S. Li and Shuang Zhou and Shao-Kang Wu and Tao Yun and Tian Pei and Tianyu Sun and T. Wang and Wangding Zeng and Wanjia Zhao and Wen Liu and Wenfeng Liang and Wenjun Gao and Wen-Xia Yu and Wentao Zhang and Wangding Xiao and Wei An and Xiaodong Liu and Xiaohan Wang and Xiaokang Chen and Xiaotao Nie and Xin Cheng and Xin Liu and Xin Xie and Xingchao Liu and Xinyu Yang and Xinyuan Li and Xuecheng Su and Xuheng Lin and X. Q. Li and Xiangyu Jin and Xi-Cheng Shen and Xiaosha Chen and Xiaowen Sun and Xiaoxiang Wang and Xinnan Song and Xinyi Zhou and Xianzu Wang and Xinxia Shan and Y. K. Li and Y. Q. Wang and Y. X. Wei and Yang Zhang and Yanhong Xu and Yao Li and Yao Zhao and Yaofeng Sun and Yaohui Wang and Yi Yu and Yichao Zhang and Yifan Shi and Yi Xiong and Ying He and Yishi Piao and Yisong Wang and Yixuan Tan and Yiyang Ma and Yiyuan Liu and Yongqiang Guo and Yuan Ou and Yuduan Wang and Yue Gong and Yu-Jing Zou and Yujia He and Yunfan Xiong and Yu-Wei Luo and Yu-mei You and Yuxuan Liu and Yuyang Zhou and Y. X. Zhu and Yanping Huang and Yao Li and Yi Zheng and Yuchen Zhu and Yunxiang Ma and Ying Tang and Yukun Zha and Yuting Yan and Zehui Ren and Zehui Ren and Zhangli Sha and Zhe Fu and Zhean Xu and Zhenda Xie and Zhen-guo Zhang and Zhewen Hao and Zhicheng Ma and Zhigang Yan and Zhiyu Wu and Zihui Gu and Zijia Zhu and Zijun Liu and Zi-An Li and Ziwei Xie and Ziyang Song and Zizheng Pan and Zhen Huang and Zhipeng Xu and Zhongyu Zhang and Zhen Zhang},
  journal={ArXiv},
  year={2025},
  volume={abs/2501.12948},
  url={https://api.semanticscholar.org/CorpusID:275789950}
}

@article{Kim2025UniversalRA,
  title={Universal Reasoner: A Single, Composable Plug-and-Play Reasoner for Frozen LLMs},
  author={Jaemin Kim and Hangeol Chang and Hyunmin Hwang and Choonghan Kim and Jong Chul Ye},
  journal={ArXiv},
  year={2025},
  volume={abs/2505.19075},
  url={https://api.semanticscholar.org/CorpusID:278904667}
}

@article{Cui2023UltraFeedbackBL,
  title={UltraFeedback: Boosting Language Models with High-quality Feedback},
  author={Ganqu Cui and Lifan Yuan and Ning Ding and Guanming Yao and Wei Zhu and Yuan Ni and Guotong Xie and Zhiyuan Liu and Maosong Sun},
  journal={ArXiv},
  year={2023},
  volume={abs/2310.01377},
  url={https://api.semanticscholar.org/CorpusID:263605623}
}

@article{Bradley1952RankAO,
  title={Rank Analysis of Incomplete Block Designs: I. The Method of Paired Comparisons},
  author={Ralph Allan Bradley and Milton E. Terry},
  journal={Biometrika},
  year={1952},
  volume={39},
  pages={324},
  url={https://api.semanticscholar.org/CorpusID:125209808}
}

@article{Zheng2025GroupSP,
  title={Group Sequence Policy Optimization},
  author={Chujie Zheng and Shixuan Liu and Mingze Li and Xionghui Chen and Bowen Yu and Chang Gao and Kai Dang and Yuqiong Liu and Rui Men and An Yang and Jingren Zhou and Junyang Lin},
  journal={ArXiv},
  year={2025},
  volume={abs/2507.18071},
  url={https://api.semanticscholar.org/CorpusID:280017753}
}

@inproceedings{Gao2022ScalingLF,
  title={Scaling Laws for Reward Model Overoptimization},
  author={Leo Gao and John Schulman and Jacob Hilton},
  booktitle={International Conference on Machine Learning},
  year={2022},
  url={https://api.semanticscholar.org/CorpusID:252992904}
}

@article{yang2025let,
  title={Let it Calm: Exploratory Annealed Decoding for Verifiable Reinforcement Learning},
  author={Yang, Chenghao and Gui, Lin and Yang, Chenxiao and Veitch, Victor and Zhang, Lizhu and Zhao, Zhuokai},
  journal={arXiv preprint arXiv:2510.05251},
  year={2025}
}

@inproceedings{Chakraborty2024TransferQS,
  title={Transfer Q Star: Principled Decoding for LLM Alignment},
  author={Souradip Chakraborty and Soumya Suvra Ghosal and Ming Yin and Dinesh Manocha and Mengdi Wang and A. S. Bedi and Furong Huang},
  year={2024},
  booktitle={The Thirty-eighth Annual Conference on Neural Information Processing Systems}
}

@article{xiong2026token,
  title={Token-Level LLM Collaboration via FusionRoute},
  author={Xiong, Nuoya and Zhou, Yuhang and Zeng, Hanqing and Chen, Zhaorun and Huang, Furong and Bi, Shuchao and Zhang, Lizhu and Zhao, Zhuokai},
  journal={arXiv preprint arXiv:2601.05106},
  year={2026}
}

@inproceedings{Huang2024DeALDA,
  title={DeAL: Decoding-time Alignment for Large Language Models},
  author={James Y. Huang and Sailik Sengupta and Daniele Bonadiman and Yi-an Lai and Arshit Gupta and Nikolaos Pappas and Saab Mansour and Katrin Kirchoff and Dan Roth},
  booktitle={Annual Meeting of the Association for Computational Linguistics},
  year={2024},
  url={https://api.semanticscholar.org/CorpusID:267616998}
}

@article{Khanov2024ARGSAA,
  title={ARGS: Alignment as Reward-Guided Search},
  author={Maxim Khanov and Jirayu Burapacheep and Yixuan Li},
  journal={ArXiv},
  year={2024},
  volume={abs/2402.01694},
  url={https://api.semanticscholar.org/CorpusID:267411977}
}

@article{Li2024CascadeRS,
  title={Cascade Reward Sampling for Efficient Decoding-Time Alignment},
  author={Bolian Li and Yifan Wang and Ananth Y. Grama and Ruqi Zhang},
  journal={ArXiv},
  year={2024},
  volume={abs/2406.16306},
  url={https://api.semanticscholar.org/CorpusID:270703542}
}

@article{Yu2025DAPOAO,
  title={DAPO: An Open-Source LLM Reinforcement Learning System at Scale},
  author={Qiying Yu and Zheng Zhang and Ruofei Zhu and Yufeng Yuan and Xiaochen Zuo and Yu Yue and Tiantian Fan and Gaohong Liu and Lingjun Liu and Xin Liu and Haibin Lin and Zhiqi Lin and Bole Ma and Guangming Sheng and Yuxuan Tong and Chi Zhang and Mofan Zhang and Wang Zhang and Hang Zhu and Jinhua Zhu and Jiaze Chen and Jiangjie Chen and Chengyi Wang and Honglin Yu and Weinan Dai and Yuxuan Song and Xiang Wei and Haodong Zhou and Jingjing Liu and Wei Ma and Ya-Qin Zhang and Lin Yan and Mu Qiao and Yong-Xu Wu and Mingxuan Wang},
  journal={ArXiv},
  year={2025},
  volume={abs/2503.14476},
  url={https://api.semanticscholar.org/CorpusID:277104124}
}

@article{Liu2025UnderstandingRT,
  title={Understanding R1-Zero-Like Training: A Critical Perspective},
  author={Zi-Yan Liu and Changyu Chen and Wenjun Li and Penghui Qi and Tianyu Pang and Chao Du and Wee Sun Lee and Min Lin},
  journal={ArXiv},
  year={2025},
  volume={abs/2503.20783},
  url={https://api.semanticscholar.org/CorpusID:277322777}
}

@article{Guha2025OpenThoughtsDR,
  title={OpenThoughts: Data Recipes for Reasoning Models},
  author={Etash Kumar Guha and Ryan Marten and Sedrick Scott Keh and Negin Raoof and Georgios Smyrnis and Hritik Bansal and Marianna Nezhurina and Jean-Pierre Mercat and Trung Vu and Zayne Sprague and Ashima Suvarna and Ben Feuer and Liangyu Chen and Zaid Khan and Eric Frankel and Sachin Grover and Caroline Choi and Niklas Muennighoff and Shiye Su and Wanjia Zhao and John Yang and Shreyas Pimpalgaonkar and Kartik Sharma and Charlie Cheng-Jie Ji and Yichuan Deng and Sarah Pratt and Vivek Ramanujan and Jon Saad-Falcon and Jeffrey Li and Achal Dave and Alon Albalak and Kushal Arora and Blake Wulfe and Chinmay Hegde and Greg Durrett and Sewoong Oh and Mohit Bansal and Saadia Gabriel and Aditya Grover and Kai-Wei Chang and Vaishaal Shankar and Aaron Gokaslan and Mike A. Merrill and Tatsunori Hashimoto and Yejin Choi and Jenia Jitsev and Reinhard Heckel and Maheswaran Sathiamoorthy and Alexandros G. Dimakis and Ludwig Schmidt},
  journal={ArXiv},
  year={2025},
  volume={abs/2506.04178},
  url={https://api.semanticscholar.org/CorpusID:279154475}
}

@article{Yue2023MAmmoTHBM,
  title={MAmmoTH: Building Math Generalist Models through Hybrid Instruction Tuning},
  author={Xiang Yue and Xingwei Qu and Ge Zhang and Yao Fu and Wenhao Huang and Huan Sun and Yu Su and Wenhu Chen},
  journal={ArXiv},
  year={2023},
  volume={abs/2309.05653},
  url={https://api.semanticscholar.org/CorpusID:261696697}
}

@article{Ye2025LIMOLI,
  title={LIMO: Less is More for Reasoning},
  author={Yixin Ye and Zhen Huang and Yang Xiao and Ethan Chern and Shijie Xia and Pengfei Liu},
  journal={ArXiv},
  year={2025},
  volume={abs/2502.03387},
  url={https://api.semanticscholar.org/CorpusID:276116748}
}

@article{Li2024Common7L,
  title={Common 7B Language Models Already Possess Strong Math Capabilities},
  author={Chen Li and Weiqi Wang and Jingcheng Hu and Yixuan Wei and Nanning Zheng and Han Hu and Zheng Zhang and Houwen Peng},
  journal={ArXiv},
  year={2024},
  volume={abs/2403.04706},
  url={https://api.semanticscholar.org/CorpusID:268264074}
}

@misc{openai2024gpt4,
  title        = {GPT-4 Technical Report},
  author       = {OpenAI},
  year         = {2024},
  howpublished = {\url{https://cdn.openai.com/papers/gpt-4.pdf}},
  note         = {Accessed: 2025-10-05}
}

@misc{deepseek-moe,
      title={DeepSeekMoE: Towards Ultimate Expert Specialization in Mixture-of-Experts Language Models}, 
      author={Damai Dai and Chengqi Deng and Chenggang Zhao and R. X. Xu and Huazuo Gao and Deli Chen and Jiashi Li and Wangding Zeng and Xingkai Yu and Y. Wu and Zhenda Xie and Y. K. Li and Panpan Huang and Fuli Luo and Chong Ruan and Zhifang Sui and Wenfeng Liang},
      year={2024},
      eprint={2401.06066},
      archivePrefix={arXiv},
      primaryClass={cs.CL},
      url={https://arxiv.org/abs/2401.06066}, 
}

@article{switch,
  author       = {William Fedus and
                  Barret Zoph and
                  Noam Shazeer},
  title        = {Switch Transformers: Scaling to Trillion Parameter Models with Simple
                  and Efficient Sparsity},
  journal      = {CoRR},
  volume       = {abs/2101.03961},
  year         = {2021},
  url          = {https://arxiv.org/abs/2101.03961},
  eprinttype    = {arXiv},
  eprint       = {2101.03961},
  bibsource    = {dblp computer science bibliography, https://dblp.org}
}

@misc{btm,
    title={Branch-Train-Merge: Embarrassingly Parallel Training of Expert Language Models},
    author={Margaret Li and Suchin Gururangan and Tim Dettmers and Mike Lewis and Tim Althoff and Noah A. Smith and Luke Zettlemoyer},
    year={2022},
    eprint={2208.03306},
    archivePrefix={arXiv},
    primaryClass={cs.CL}
}

@misc{btx,
    title={Branch-Train-MiX: Mixing Expert LLMs into a Mixture-of-Experts LLM},
    author={Sainbayar Sukhbaatar and Olga Golovneva and Vasu Sharma and Hu Xu and Xi Victoria Lin and Baptiste Rozière and Jacob Kahn and Daniel Li and Wen-tau Yih and Jason Weston and Xian Li},
    year={2024},
    eprint={2403.07816},
    archivePrefix={arXiv},
    primaryClass={cs.CL}
}

@misc{mixlora,
      title={MixLoRA: Enhancing Large Language Models Fine-Tuning with LoRA-based Mixture of Experts}, 
      author={Dengchun Li and Yingzi Ma and Naizheng Wang and Zhengmao Ye and Zhiyuan Cheng and Yinghao Tang and Yan Zhang and Lei Duan and Jie Zuo and Cal Yang and Mingjie Tang},
      year={2024},
      eprint={2404.15159},
      archivePrefix={arXiv},
      primaryClass={cs.CL},
      url={https://arxiv.org/abs/2404.15159}, 
}

@misc{hydralora,
      title={HydraLoRA: An Asymmetric LoRA Architecture for Efficient Fine-Tuning}, 
      author={Chunlin Tian and Zhan Shi and Zhijiang Guo and Li Li and Chengzhong Xu},
      year={2024},
      eprint={2404.19245},
      archivePrefix={arXiv},
      primaryClass={cs.CL},
      url={https://arxiv.org/abs/2404.19245}, 
}

@misc{smore,
      title={S'MoRE: Structural Mixture of Residual Experts for LLM Fine-tuning}, 
      author={Hanqing Zeng and Yinglong Xia and Zhuokai Zhao and Gilbert Jiang and Qiang Zhang and Jiayi Liu and Lizhu Zhang and Xiangjun Fan and Benyu Zhang},
      year={2025},
      eprint={2504.06426},
      archivePrefix={arXiv},
      primaryClass={cs.CL},
      url={https://arxiv.org/abs/2504.06426}, 
}

\appendix
\onecolumn
\section*{Appendix}
The appendix provides supporting details across eight sections. \appref{app:exp_details} describes implementation and hyperparameter details for training the reward model and token-level router. \appref{app:router_complexity} analyzes the parameter overhead introduced by the router for each design variant. \appref{app:router_inputs} extends the router input ablation to all three benchmarks. \appref{app:majority_voting} compares TARo against majority voting on Qwen2.5-3B. \appref{app:token_vs_prompt} validates the choice of token-level over prompt-level routing. We provide generation and evaluation (LLM-as-a-judge) prompts in \appref{app:gen-prompt} and \appref{app:eval-prompt}, respectively. Finally, \appref{appx:addiotion-qual-ex} presents additional qualitative examples illustrating cases where TARo succeeds and where it fails.
\section{Experiments Implementation and Hyperparameter Details}
\label{app:exp_details}

\paragraph{Reasoning reward model.}
Reward model training uses AdamW with a learning rate of $2\times 10^{-5}$, cosine learning rate scheduler, batch size 32, and 3 epochs. We set $\beta_r = 0.1$ in the preference loss. LoRA adapters are applied with rank $8$ and scaling factor $\alpha=16$ \footnote{This $\alpha$ differs from the interpolation coefficient in GenARM~\cite{Xu2024GenARMRG} and our router}. 

\paragraph{Token-level router.}
Router training uses supervised fine-tuning (SFT) to train the MLP layers (hidden dimension size $H=128$) on 1,000 samples from each dataset for three epochs. We use a learning rate of $5\times 10^{-6}$, batch size 32, and $10$ warmup steps. Training is conducted on 1-4 NVIDIA H100 GPUs with bfloat16 precision.

\section{Router Complexity}
\label{app:router_complexity}
Compared with GenARM, the router introduces $(2V+1)\times(H+1)$ learnable parameters for Design (i) and $(2Kd+1)\times(H+1)+d\times V$ for Design (ii). Given the hidden dimension of router prediction head $H=128$, for Llama ($V=128256$) and Qwen 2.5 ($V=151936$), the router would have $33M$ and $12M$ learnable parameters respectively.

\section{Router Input Ablation: Full Results}
\label{app:router_inputs}
\begin{table}[h]
\centering
\small
\begin{tabular}{lccccc}
\toprule
\textbf{Router Inputs} & \textbf{Scalable to} & \textbf{MATH500} & \textbf{AlpacaEval} & \textbf{MedXpertQA} \\
 & \textbf{Larger Base} & \textbf{(Acc \%)} & \textbf{(LC WR \%)} & \textbf{(Acc \%)} & \textbf{Avg.} \\ 
\midrule
\rowcolor{lightgray}
Base + reward hidden states & \xmark & 49.6 & 18.7 & \textbf{13.7} & 27.3\\
Reward hidden state only    & \cmark & 51.2 & 15.7 & 12.6 & 26.5 \\
Reward logits only & \cmark & 52.6 & 19.7 & \textbf{14.4} & 28.9 \\
Base + reward logits        & \cmark & \textbf{54.4} & \textbf{20.8} & 13.2 & \textbf{29.5} \\
\bottomrule
\end{tabular}
\caption{Ablation of router input configurations on Llama3.1-8B across all three benchmarks. \textit{Scalable to Larger Base} indicates whether the router can be 
applied to a larger base model without retraining. The greyed row is excluded from 
the scalable comparison. Among scalable designs, base and reward logits 
on average outperforms using hidden states or logits from the reward model alone. Bold indicates best result per column across all rows.}
\label{tab:router_inputs_full}
\end{table}

\tabref{tab:router_inputs_full} extends the router input ablation from the main paper (\secref{subsec:ablations_on_token-level_router}) to all three benchmarks. We first compare two router input configurations: using reward hidden states alone vs. using both base and reward hidden states. The latter consistently improves performance on tasks outside the reward model's training domain (MedXpertQA: $12.6\%\to13.7\%$, AlpacaEval: $15.7\%
\to18.7\%)$, suggesting that the reward model's hidden states carry a domain bias from mathematical training that the base model's representation can counteract. However, this configuration is not base-model-scalable, as a larger base model's hidden states may differ in both representation and dimensionality from those seen during router training.

 Logits provide a scale-agnostic alternative, where the vocabulary space remains the same across scales. We observe that reward logits alone already generalize better on OOD tasks than their hidden state counterpart (AlpacaEval: 19.7\% vs. 15.7\%, MedXpertQA: 14.4\% vs.\ 12.6\%). We posit this is because logits encode model confidence over a shared, domain-agnostic vocabulary space, making them a more universal signal for routing regardless of the target domain. Adding base logits further improves in-domain performance (MATH500: 52.6\%$\to$54.4\%) at a slight OOD cost, and achieves the best average across benchmarks, motivating our choice of base and reward logits as the default configuration.

\section{Comparison with Majority Voting}
\label{app:majority_voting}

\begin{table}[h]
\centering
\small
\begin{tabular}{lcccc}
\toprule
\textbf{Method} & \textbf{MATH500} & \textbf{AlpacaEval} & \textbf{MedXpertQA} & \textbf{Est. TFLOPs} \\
 & \textbf{(Acc \%)} & \textbf{(LC WR \%)} & \textbf{(Acc \%)} & \\
\midrule
Majority Voting ($N$=8) & 56.4 & --- & \textbf{10.6} & 113.6 \\
GenARM                  & 56.4 & 18.9 & 9.8  & 28.4  \\
TARo                    & \textbf{64.8} & \textbf{21.6} & 9.8  & 28.4  \\
\bottomrule
\end{tabular}
\caption{Comparison with majority voting on Qwen2.5-3B. TFLOPs are estimated 
for a single forward pass with a 2048-token context. Majority voting is not applicable to AlpacaEval (---) due to its open-ended 
generation format. Bold indicates best result per column.}
\label{tab:majority_voting}
\end{table}

\tabref{tab:majority_voting} compares TARo against majority voting and GenARM 
on Qwen2.5-3B. Note that majority voting cannot be applied to AlpacaEval, as it depends on a known, extractable answer format. This can make majority voting poorly suited for agentic systems, as a potential application setting, that need flexible logical reasoning and instruction-following, not fixed-format outputs. TARo outperforms majority voting on MATH500 while requiring only 
$4\times$ less compute, as majority voting requires $N{=}8$ full response samples rather than token-wise generation. On MedXpertQA, TARo underperforms majority 
voting. 

\section{Prompt-level vs. Token-level Routing}
\label{app:token_vs_prompt}

\begin{table}[h]
\centering
\small
\begin{tabular}{lcccc}
\toprule
\textbf{Routing Granularity} & \textbf{MATH500} & \textbf{AlpacaEval} & \textbf{MedXpertQA} & \textbf{Avg.} \\
 & \textbf{(Acc \%)} & \textbf{(LC WR \%)} & \textbf{(Acc \%)} \\
\midrule
Prompt-level & 33.2 & \textbf{22.1} & 13.3 & \\
Token-level  & \textbf{49.6} & 18.7 & \textbf{13.7} & \\
\bottomrule
\end{tabular}
\caption{Comparison of prompt-level versus token-level routing with hidden state inputs on Llama3.1-8B. Token-level routing 
significantly outperforms prompt-level routing on mathematical reasoning, 
validating the design choice of adaptive token-level $\alpha$ prediction.}
\label{tab:token_vs_prompt}
\end{table}

\tabref{tab:token_vs_prompt} compares router performance when trained to predict a single prompt-level $\alpha$ (uniform across all tokens) versus a token-level $\alpha$ at each decoding step. Prompt-level routing performs significantly worse on mathematical reasoning (MATH500: 33.2\% vs.\ 49.6\%), demonstrating that fine-grained token-level control is essential for structured reasoning tasks where the reward model's guidance should be concentrated on specific tokens such as operators, variables, and reasoning scaffolds (see Tab.~\ref{tab:alpha-top} from main text). Performance differences on OOD tasks (AlpacaEval, MedXpertQA) are smaller, suggesting that token-level granularity matters most in the in-domain reasoning setting.

\section{Generation Prompts}
\label{app:gen-prompt}

We share the MATH500 and MedXpertQA generation prompts in this section.
\begin{tcolorbox}[promptbox, title=MATH500 Generation Prompt]
\begin{lstlisting}[breaklines,basicstyle=\normalfont\scriptsize,mathescape=true]

Solve the following math problem step by step. Please reason step by step, and put your final answer within \texttt{\textbackslash boxed\{\}}.

Here are some examples:

Problem:
Find the domain of the expression $\frac{\sqrt{x-2}}{\sqrt{5-x}}$.

Solution:
The expressions inside each square root must be non-negative.
Therefore, $x-2 \ge 0$, so $x\ge2$, and $5 - x \ge 0$, so $x \le 5$.
Also, the denominator cannot be equal to zero, so $5-x>0$, which gives $x<5$.
Therefore, the domain of the expression is $\boxed{[2,5)}$.

Final Answer: The final answer is $[2,5)$. I hope it is correct.

Problem:
If $\det \mathbf{A} = 2$ and $\det \mathbf{B} = 12,$ then find $\det (\mathbf{A} \mathbf{B}).$

Solution:
We have that $\det (\mathbf{A} \mathbf{B}) = (\det \mathbf{A})(\det \mathbf{B}) = (2)(12) = \boxed{24}.$

Final Answer: The final answer is $24$. I hope it is correct.

Problem:
Terrell usually lifts two 20-pound weights 12 times. If he uses two 15-pound weights instead, how many times must Terrell lift them in order to lift the same total weight?

Solution:
If Terrell lifts two 20-pound weights 12 times, he lifts a total of $2\cdot 12\cdot20=480$ pounds of weight.  If he lifts two 15-pound weights instead for $n$ times, he will lift a total of $2\cdot15\cdot n=30n$ pounds of weight.  Equating this to 480 pounds, we can solve for $n$: \begin{align*}
30n&=480\\
\Rightarrow\qquad n&=480/30=\boxed{16}
\end{align*}

Final Answer: The final answer is $16$. I hope it is correct.

Problem:
If the system of equations

\begin{align*}
6x-4y&=a,\\
6y-9x &=b.
\end{align*}has a solution $(x, y)$ where $x$ and $y$ are both nonzero, find $\frac{a}{b},$ assuming $b$ is nonzero.

Solution:
If we multiply the first equation by $-\frac{3}{2}$, we obtain

$$6y-9x=-\frac{3}{2}a.$$Since we also know that $6y-9x=b$, we have

$$-\frac{3}{2}a=b\Rightarrow\frac{a}{b}=\boxed{-\frac{2}{3}}.$$

Final Answer: The final answer is $-\frac{2}{3}$. I hope it is correct.
\end{lstlisting}
\end{tcolorbox}

\clearpage

\begin{tcolorbox}[promptbox, title=MedXpertQA Generation Prompt]
\begin{lstlisting}[breaklines,basicstyle=\normalfont\scriptsize,mathescape=true]

# Instructions
* Select the multiple choice answer that best answers the question.
* Please think step by step.
* Write your final answer as "Selected Answer Choice: {answer}"

# Examples

User: A 29-year-old woman presents to the emergency department with a 1-day history of fever. She says that it started spontaneously and that it hasn't gotten any better with either rest or acetaminophen. Her past medical history is significant for homelessness, unprotected sexual intercourse with multiple partners, and substance use disorder with multiple substances. Her temperature is 103.5F (39.5C), blood pressure is 100/70 mmHg, pulse is 112/min, respirations are 18/min, and oxygen saturation is 93% on room air. On physical exam, a murmur is heard along the left mid-sternal border. The pulmonary exam reveals minor bibasilar crackles. The patient's upper limbs demonstrate many bruises and scars in the antecubital fossa. The patient is started on vancomycin and gentamicin and sent for echocardiography. Based on the results of the echocardiogram, the patient undergoes surgery to remove multiple vegetations from the tricuspid valve. Vancomycin and gentamicin are continued over the next 5 days. On postoperative day 5, the patient presents with bleeding from her gums, oozing from her surgical sites, and recurrent epistaxis. Lab values are obtained as seen below:
Serum:Na+: 135 mEq/L
Cl-: 90 mEq/L
K+: 4.4 mEq/L
HCO3-: 23 mEq/L
BUN: 20 mg/dL
Glucose: 110 mg/dL
Creatinine: 1.0 mg/dL
Ca2+: 10.1 mg/dL
AST: 9 U/L
ALT: 9 U/L
Leukocyte count and differential:
Leukocyte count: 6,000 cells/mm^3
Lymphocytes: 20%
Monocytes: 1%
Neutrophils: 78%
Eosinophils: 1%
Basophils: 0%
PT: 27 seconds
aPTT: 84 seconds
D-dimer: < 50 $\mu$g/L
Hemoglobin: 14 g/dL
Hematocrit: 40%
Platelet count: 150,000/mm^3
Mean corpuscular volume: 110 fL
Mean corpuscular hemoglobin concentration: 34 g/dL
RDW: 14%
Which of the following is the most likely cause of this patient's current symptoms?
Answer Choices: (A) Bacterial infection of the bloodstream (B) Defect in von Willebrand factor (C) Acquired hemophilia A (D) Factor VIII deficiency (E) Coagulation cascade activation (F) Disseminated intravascular coagulation (G) Vitamin K deficiency due to antibiotic use (H) Immune-mediated thrombocytopenia (I) Antibiotic therapy (J) Hypersensitivity reaction to antibiotics
<|eot_id|>

Assistant: To determine the most likely cause of this patient's current symptoms, let's analyze the situation step by step. A 29-year-old woman presents to the emergency department with a 1-day history of fever. She has a history of homelessness, unprotected sexual intercourse with multiple partners, and substance use disorder. Her vital signs show a high fever, tachycardia, and slightly low oxygen saturation. A murmur is heard along the left mid-sternal border, and there are minor bibasilar crackles on pulmonary exam. The presence of bruises and scars in the antecubital fossa suggests intravenous drug use. She is started on vancomycin and gentamicin and undergoes surgery to remove vegetations from the tricuspid valve, indicating infective endocarditis. On postoperative day 5, she presents with bleeding from her gums, oozing from surgical sites, and recurrent epistaxis. Lab values show prolonged PT and aPTT, normal platelet count, and normal D-dimer, which suggests a coagulopathy rather than thrombocytopenia or DIC. The prolonged PT and aPTT with normal platelet count and D-dimer are consistent with a coagulation factor deficiency or inhibition. Given the recent antibiotic use, vitamin K deficiency due to antibiotic therapy is a likely cause, as antibiotics can disrupt gut flora that synthesize vitamin K, leading to deficiencies in vitamin K-dependent clotting factors. Therefore, the most likely cause of this patient's current symptoms is: Antibiotic therapy.

Selected Answer Choice: I
<|eot_id|>

User: A 55-year-old man is brought to the emergency department after a witnessed tonic-clonic seizure. There was no trauma during the seizure. The patient's medical history includes hypertension, coronary artery disease, and type 2 diabetes. His current medications are hydrochlorothiazide, atenolol, metformin, and aspirin. According to his family, he smokes a pack of cigarettes and consumes 6 to 8 beers daily but does not use recreational drugs. His last alcohol consumption occurred approximately 36 hours before presentation.

His blood pressure is 160/110 mm Hg, his heart rate is 98 beats per minute and regular, and his respiratory rate is 22 breaths per minute. He is afebrile but confused and diaphoretic.

A complete blood count shows mild macrocytic anemia and thrombocytopenia. A biochemical profile shows a normal glucose level, mild hypokalemia, hypophosphatemia, normal kidney function, and an elevated aspartate aminotransferase level.

Chest radiography is unremarkable. Noncontrast CT of the brain shows mild volume loss. Blood and urine drug test results are negative. An electrocardiogram is notable for sinus tachycardia and motion artifact.

Thiamine, multivitamins, folate, and electrolyte repletion are administered. Three hours later, the patient becomes increasingly agitated and confused, with incoherent speech, diaphoresis, and combative behavior that persist even after the cumulative administration of 40 mg of intravenous lorazepam. Physical restraints to prevent him from disrupting his intravenous access only contribute to his agitation.

What is the most appropriate initial management option in this patient?
Answer Choices: (A) Obtain repeat CT of the brain (B) Transfer to the intensive care unit, and initiate sedation (C) Start a continuous infusion of midazolam (D) Obtain electroencephalography (E) Start a continuous infusion of phenytoin (F) Administer phenobarbital for refractory agitation (G) Perform lumbar puncture to rule out meningitis (H) Administer haloperidol for agitation (I) Initiate propofol infusion for sedation (J) Administer dexmedetomidine for sedation
<|eot_id|>

Assistant: To determine the most appropriate initial management option for this patient, let's analyze the situation step by step. The patient is a 29-year-old woman with a history of homelessness, unprotected sexual intercourse with multiple partners, and substance use disorder. She presents with a high fever, tachycardia, and slightly low oxygen saturation. Physical examination reveals a murmur along the left mid-sternal border and minor bibasilar crackles. The presence of bruises and scars in the antecubital fossa suggests intravenous drug use. She is started on vancomycin and gentamicin due to suspected infective endocarditis and undergoes surgery to remove vegetations from the tricuspid valve. On postoperative day 5, she presents with bleeding from her gums, oozing from surgical sites, and recurrent epistaxis. Lab values show prolonged PT and aPTT, normal platelet count, and normal D-dimer, indicating a coagulopathy rather than thrombocytopenia or disseminated intravascular coagulation (DIC). The prolonged PT and aPTT with normal platelet count and D-dimer suggest a coagulation factor deficiency or inhibition. Given the recent antibiotic use, vitamin K deficiency due to antibiotic therapy is a likely cause, as antibiotics can disrupt gut flora that synthesize vitamin K, leading to deficiencies in vitamin K-dependent clotting factors. The most likely cause of this patient's current symptoms is antibiotic therapy.

Selected Answer Choice: B
<|eot_id|>
\end{lstlisting}
\end{tcolorbox}

\newpage
\section{Evaluation prompts.}
\label{app:eval-prompt}
In this section, we share the evaluation prompt used by \texttt{Llama3.1-70B} to evaluate the extracted answer for \texttt{MedXpertQA} and \texttt{MATH500}. 
First, we extract the model’s answer from the generated output by (1) taking the content inside ``\texttt{\textbackslash boxed\{\}}'' for Qwen, 
(2) taking the span after ``Final answer:'' for Llama, or (3) using the full generated output as a fallback for either if extraction based on formatting fails. The evaluation prompt considers equivalence between the extracted answer and the ground truth answer for MedXpertQA. For MATH500, the evaluation prompt provides examples to evaluate based on mathematical equivalence.

\begin{tcolorbox}[promptbox, title=MATH500 Evaluation Prompt]
\begin{lstlisting}[breaklines,basicstyle=\normalfont\scriptsize]
Look at the following two expressions (answers to a math problem) and judge whether they are equivalent. Only perform trivial simplifications

Examples:

    Expression 1: $2x+3$
    Expression 2: $3+2x$

Yes

    Expression 1: 3/2
    Expression 2: 1.5

Yes

    Expression 1: $x^2+2x+1$
    Expression 2: $y^2+2y+1$

No

    Expression 1: $x^2+2x+1$
    Expression 2: $(x+1)^2$

Yes

    Expression 1: 3245/5
    Expression 2: 649

No
(these are actually equal, don't mark them equivalent if you need to do nontrivial simplifications)

    Expression 1: 2/(-3)
    Expression 2: -2/3

Yes
(trivial simplifications are allowed)

    Expression 1: 72 degrees
    Expression 2: 72

Yes
(give benefit of the doubt to units)

    Expression 1: 64
    Expression 2: 64 square feet

Yes
(give benefit of the doubt to units)

---

YOUR TASK


Respond with only "Yes" or "No" (without quotes). Do not include a rationale.

    Expression 1: %(expression1)s
    Expression 2: %(expression2)s
\end{lstlisting}
\end{tcolorbox}

\clearpage
\begin{tcolorbox}[promptbox, title=MedXpertQA Evaluation Prompt]
\begin{lstlisting}[breaklines,basicstyle=\scriptsize]
Look at the following two answers and judge whether they are equivalent.
---
YOUR TASK


Respond with only "Yes" or "No" (without quotes). Do not include a rationale.

    Answer 1: %(expression1)s
    Answer 2: %(expression2)s
\end{lstlisting}
\end{tcolorbox}

\clearpage


\newpage
\section{Additional Qualitative Examples}
\label{appx:addiotion-qual-ex}
We present additional qualitative examples comparing the base, reasoning reward, and \algname model outputs in this section. Table \ref{tab:model-comparison-base-fails} illustrates a case where the base model bypasses the reasoning process, producing an incorrect result (200), while both the reward and guided models reason systematically and arrive at the correct answer (2220). In contrast, Table \ref{tab:model-comparison-guide-reward-fails} presents a case where the base model quickly identifies the correct cube (27) through trial and error, but the reward and \algname follow the right strategy, but make an error during reasoning, leading to an incorrect answer. This suggests that the reward and guided model lacks mechanisms to revise their problem-solving strategy and check their work. Future work can enhance the reward model's ability to reward backtracking and verification.

\begin{table*}[h]
\small
  \centering
  \begin{tabular}{p{1.4cm}p{6cm}p{3.5cm}p{3.5cm}} \hline
      \multicolumn{4}{p{\linewidth}}{\textbf{Question:} What is the least positive integer multiple of 30 that can be written with only the digits 0 and 2?} \\
      \multicolumn{4}{p{\linewidth}}{\textbf{Ground Truth:} 2220} \\
    \hline
     & \textbf{Base Model} & \textbf{Reward Model} & \textbf{\algname (Ours)} \\
    \hline
    \textbf{Output} &
The least positive integer multiple of 30 that can be written with only the digits 0 and 2 is $\boxed{200}$.  
Final Answer: 200. I hope it is correct. &
We need to find the smallest positive integer multiple of 30 that can be written with only the digits 0 and 2. Since 30 is a multiple of 2, 3, and 5, the number must end with a 0 (to be divisible by 10) and contain at least one 3 or 9 (to be divisible by 3). However, we are only allowed to use the digits 0 and 2, so the number must be written with only the digits 0 and 2. The smallest such number is 200, but 200 is not divisible by 3. The next possible number is 2220, which is divisible by 6 (and therefore by 3) and ends with a 0 (so it is divisible by 10). Therefore, 2220 is a multiple of 30. The smallest such number is 60, but 60 is not written with only the digits 0 and 2. The next possible number is 22020, but 22020 is not written with only the digits 0 and 2. The smallest such number is 2220, which is written with only the digits 0 and 2 and is a multiple of 30. Therefore, the answer is 2220.  
Final Answer: 2220. I hope it is correct. &
\#\# Step 1: Understand the problem  
We need to find the smallest multiple of 30 that can be expressed using only the digits 0 and 2.  

\#\# Step 2: Recall divisibility rules  
Multiples of 30 are divisible by both 2 and 3, and must end with 0.  

\#\# Step 3: Divisibility by 3  
The sum of digits must be a multiple of 3. With digits 0 and 2, the sum = 2×(number of 2s), so the number of 2s must be a multiple of 3.  

\#\# Step 4: Try smallest combination  
Three 2s and one 0 gives 2220.  

\#\# Step 5: Verification  
2220 is divisible by 2, 3, and 5.  
$\Rightarrow$ It is divisible by 30.  

Final Answer: $\boxed{2220}$ \\
    \hline
  \end{tabular}
  \caption{Qualitative example comparing \textbf{Llama-3.1 8B Instruct} (base), the \textbf{reasoning reward model}, and \algname on a divisibility problem. The base model bypasses reasoning, while both the reward and guided models reason systematically to reach the correct answer.}
  \label{tab:model-comparison-base-fails}
\end{table*}
\begin{table*}[h]
\small
  \centering
  \begin{tabular}{p{1.0cm}p{4cm}p{2.5cm}p{7cm}} \hline
      \multicolumn{4}{p{\linewidth}}{\textbf{Question:} What is the smallest positive perfect cube that can be written as the sum of three consecutive integers?} \\
      \multicolumn{4}{p{\linewidth}}{\textbf{Ground Truth:} 27} \\
    \hline
     & \textbf{Base Model} & \textbf{Reward Model} & \textbf{\algname (Ours)} \\
    \hline
    \textbf{Output} &
We can check small sets of three consecutive integers: $\{1,2,3\}, \{2,3,4\},
\{3,4,5\}, \ldots$. The sums are 6, 9, 12, 15, 18, 21, 24, 27. The first perfect cube among these is 27, which corresponds to $\{8,9,10\}$.  
Final Answer: $\boxed{27}$. &
We need to find the smallest positive perfect cube that can be written as the sum of three consecutive integers. Let's denote the three consecutive integers as $n-1$, $n$, and $n+1$. Then, the sum of these three integers is $(n-1)+n+(n+1)=3n$. We need to find the smallest positive perfect cube that can be written as $3n$. Let's consider the cubes of the positive integers and see if any of them can be written as the sum of three consecutive integers. For example, the cube of 1 is 1, the cube of 2 is 8, the cube of 3 is 27, the cube of 4 is 64, the cube of 5 is 125, and the cube of 6 is 216. The cube of 6 is 216. Let's check if 216 can be written as the sum of three consecutive integers. We can write 216 as 3n, so $n=72$. Then, the three consecutive integers are 71, 72, and 73. The sum of these three integers is $71+72+73=216$. Therefore, the smallest positive perfect cube that can be written as the sum of three consecutive integers is 216. Final Answer: The final answer is 216. I hope it is correct. &
Let's take a look at the problem. We're looking for the smallest positive perfect cube that can be written as the sum of three consecutive integers. Let's start by finding the smallest perfect cube. The smallest positive perfect cube is $1^3=1$. Now, we need to find three consecutive integers that add up to 1. The only set of three consecutive integers that add up to 1 is 0, 1, and 2, but their sum is 3, not 1. Let's try the next perfect cube. The next smallest positive perfect cube is $2^3=8$. We need to find three consecutive integers that add up to 8. The set of three consecutive integers that add up to 8 is -3, -2, and -1. Their sum is -6, not 8. The next perfect cube is $3^3=27$. We need to find three consecutive integers that add up to 27. The set of three consecutive integers that add up to 27 is -10, -9, and -8. Their sum is -27, not 27. The next perfect cube is $4^3=64$. We need to find three consecutive integers that add up to 64. The set of three consecutive integers that add up to 64 is -21, -20, and -19. Their sum is -60, not 64. The next perfect cube is $5^3=125$. We need to find three consecutive integers that add up to 125. The set of three consecutive integers that add up to 125 is -44, -43, and -42. Their sum is -129, not 125. The next perfect cube is $6^3=216$. We need to find three consecutive integers that add up to 216. The set of three consecutive integers that add up to 216 is -71, -70, and -69. Their sum is -210, not 216. The next perfect cube is $7^3=343$. We need to find three consecutive integers that add up to 343. The set of three consecutive integers that add up to 343 is -112, -111, and -110. Their sum is -333, not 343. The next perfect cube is $8^3=512$. We need to find three consecutive integers that add up to 512. The set of three consecutive integers that add up to 512 is -169, -168, and -167. Their sum is -504, not 512. The next perfect cube is $9^3=729$. We need to find three consecutive integers that add up to 729. The set of three consecutive integers that add up to 729 is -242, -241, and -240. Their sum is -723, not 729. The next perfect cube is $10^3=1000$. We need to find three consecutive integers that add up to 1000. The set of three consecutive integers that add up to 1000 is -333, -332, and -331. Their sum is -996, not 1000. The next perfect cube is $11^3=1331$. We need to find three consecutive integers that add up to 1331. The set of three consecutive integers that add up to 1331 is -444, -443, and -442. Their sum is -1329, not 1331. The next perfect cube is $12^3=1728$. We need to find three consecutive integers that add up to 1728. The set of three consecutive integers that add up to 1728 is -579, -578, and -577. Their sum is -1734, not 1728. The next perfect cube is $13^3=2197$. We need to find three consecutive integers that add up to 2197. The set of three consecutive integers that add up to 2197 is -728, -727, and -726. Their sum is -2181, not 2197. The next perfect cube is $14^3=2744$. We need to find three consecutive integers that add up to 2744. The set of three consecutive integers that add up to 2744 is -913, -912, and -911. Their sum is -2736, not 2744. The next perfect cube is $15^3=3375$. We need to find three consecutive integers that add up to 3375. The set of three consecutive integers that add up to 3375 is -1126, -1125, and -1124. Their sum is -3375, which is 3375. Therefore, the smallest positive perfect cube that can be written as the sum of three consecutive integers is $3375$.
The final answer is: $\boxed{3375}$ \\
    \hline
  \end{tabular}
  \caption{Qualitative example comparing \textbf{Llama-3.1 8B Instruct} (base), the \textbf{reasoning reward model}, and \algname on a simple cube-sum problem. The base model correctly identifies the answer through simple enumeration strategy reasoning, while the reward and guided models added errors during reasoning, leading to incorrect answers.}
  \label{tab:model-comparison-guide-reward-fails}
\end{table*}

\end{document}